\documentclass[journal]{IEEEtran}  

\IEEEoverridecommandlockouts                             

\usepackage{amsmath} 
\usepackage{graphicx}
\graphicspath{{./fig/}}
\usepackage{algorithm}
\usepackage{algorithmic}
\usepackage{mathtools}
\usepackage{hyperref}
\usepackage{multicol}
\usepackage{caption}
\usepackage{subcaption}
\usepackage{amssymb}
\usepackage{siunitx}
\usepackage{cuted}
\usepackage{lipsum}
\usepackage{multirow}
\usepackage{booktabs}
\usepackage[dvipsnames]{xcolor}

\DeclarePairedDelimiter\abs{\lvert}{\rvert}%

\title{\LARGE \bf
Robust walking based on MPC with viability guarantees*
}

\author{Mohammad Hasan Yeganegi$^{1}$, Majid Khadiv$^{1}$, Andrea Del Prete$^{2}$,
S. Ali A. Moosavian$^{3}$, Ludovic Righetti$^{1,4}$
\thanks{*This work is supported by New York University, the European Union's Horizon 2020 research and innovation program (grant agreement No 780684) and the National Science Foundation (grant CMMI-1825993)}
\thanks{$^{1}$ Max Planck Institute for Intelligent Systems, Tuebingen, Germany. {\tt\small firstname.lastname@tuebingen.mpg.de}}%
\thanks{$^{2}$ Industrial Engineering Department, University of
Trento, Italy. {\tt\small andrea.delprete@unitn.it}}
\thanks{$^{3}$ K. N. Toosi University of Technology, Tehran, Iran.  {\tt\small moosavian@kntu.ac.ir}}%
\thanks{$^{4}$ Tandon School of Engineering, New York University, New York, USA. {\tt\small ludovic.righetti@nyu.edu}}%
}

\newtheorem{remark}{\textbf{Remark}}
\begin{document}

\maketitle
\thispagestyle{empty}
\pagestyle{empty}

\begin{abstract}

Model predictive control (MPC) has shown great success for controlling complex systems such as legged robots. However, when closing the loop, the performance and feasibility of the finite horizon optimal control problem (OCP) solved at each control cycle is not guaranteed anymore. This is due to model discrepancies, the effect of low-level controllers, uncertainties and sensor noise. To address these issues, we propose a modified version of a standard MPC approach used in legged locomotion with viability (weak forward invariance) guarantees. In this approach, instead of adding a (conservative) terminal constraint to the problem, we propose to use the measured state projected to the viability kernel in the OCP solved at each control cycle. Moreover, we 
use past experimental data to find the best cost weights, which measure a combination of performance, constraint satisfaction robustness, or stability (invariance). These interpretable costs measure the trade off between robustness and performance. For this purpose, we use Bayesian optimization (BO) to systematically design experiments that help efficiently collect data to learn a cost function leading to robust performance. Our simulation results with different realistic disturbances (i.e. external pushes, unmodeled actuator dynamics and computational delay) show the effectiveness of our approach to create robust controllers for humanoid robots.

\end{abstract}

\section{INTRODUCTION}
Thanks to the increase of computational power, employing model predictive control (MPC) to control complex mechanical systems such as legged robots is feasible nowadays. Using MPC for controlling legged robots is desirable, because 1) these robots are expected to perform tasks in dynamically changing environments 2) they have highly limiting interaction constraints and 3) MPC affords the prediction of steps in the future as a response to current disturbances.

Although MPC has been shown to be a successful paradigm for controlling legged robots \cite{herdt2010online,audren2014model,grandia2019feedback}, dealing with uncertainties in the underlying optimal control problem (OCP) is only tractable for simplified cases \cite{villa2017model,gazar2020stochastic}. Furthermore, it is crucial to add a terminal constraint to the problem to guarantee the invariance of the finite horizon OCP, in which case ensuring feasibility of this constrained optimization problem becomes an issue rarely addressed in the field  \cite{villa2017model}. It is however important to consider this problem because the dynamic model used for MPC does not take into account the true dynamics of the robot nor the effect of the low-level controllers and it rarely models sensor noise and environmental uncertainty. As a consequence, merely using the measured state of the robot in the OCP can lead to an infeasible problem when a conservative terminal constraint is considered, or cause divergence of the center of mass (CoM) motion when no terminal constraint is imposed, despite the robot remaining capable to maintain balance.
Additionally, it becomes very difficult to find a cost function for the OCP that improves robustness and performance in face of all these unmodeled effects.

In this paper, we tackle the problems of invariance of the desired CoM trajectory and constraint satisfaction robustness of constrained MPC for humanoid walking. 
We propose a novel approach to adapt the estimated current state of the system used in the constrained MPC to prevent the desired CoM trajectory from divergence at all times. 
We use the viability kernel bounds to compute a feasible state while minimizing departure from the measured state of the system.
Furthermore, 
we propose to adapt the cost function of the OCP to increase the robustness of the controller to unknown dynamics and environmental uncertainty and to implicitly take into account the, possibly complex, low-level controllers. We aim to find this cost function in as few experiments as possible using Bayesian Optimization (BO).
We demonstrate the capabilities of our approach in a complete system that also includes a complex low-level model predictive controller and realistic uncertainties in a full-body simulation. 

\subsection{Related work}
\subsubsection{MPC for locomotion}
One of the earliest works that employed MPC for controlling legged robots is \cite{wieber2006trajectory}. In this work, Wieber modified the formulation in \cite{kajita2003biped} and introduced MPC as a strong tool for controlling walking of highly constrained and inherently unstable legged robots. After this work, MPC has become one of the dominant approaches for controlling legged robots. Apart from walking, where linear MPC can be implemented thanks to the linear inverted pendulum model (LIPM) \cite{herdt2010online,khadiv2020walking}, there has been a tremendous effort in the community to make this feasible for more complex models and tasks \cite{dai2016planning,carpentier2018multicontact,ponton2020efficient}.

All of the aforementioned approaches use a deterministic representation of the problem and rely on the low-level feedback control and the inherent robustness of MPC for dealing with disturbances and uncertainties. To make the controller robust, \cite{del2016robustness} took the effects of uncertainties into account and proposed a robust approach to constraint satisfaction in the low-level instantaneous feedback controller. \cite{dai2016planning,manchester2019robust,yeganegi2019robust} addressed the problem of \emph{robust trajectory optimization}, where the goal is to generate trajectories that are far from the boundaries of constraints (constraint satisfaction robustness). 
Although similar in constraint satisfaction aspect, tackling robustness in MPC problems introduces two more issues with respect to the robust trajectory optimization problem. First, since it is not clear what the final cost/constraints of the locomotion problem are, ensuring invariance properties or stability of the system is very challenging. Second, when a (conservative) terminal constraint is taken into account, the finite horizon OCP can easily become infeasible in the presence of uncertainties.

Recently, uncertainties have been considered in the MPC synthesis using robust (RMPC) \cite{villa2017model} and stochastic (SMPC) \cite{gazar2020stochastic} approaches. Using the measured state of the system directly inside MPC problem may render the OCP infeasible (in the presence of any state constraint). This is a well-known problem of constrained MPC in general and if the OCP is feasible at all time instants given that it is feasible at initial time, it is said to be \emph{recursively feasible} \cite{Kouvaritakis2016}.
To ensure recursive feasibility, \cite{villa2017model} employed the RMPC approach in \cite{mayne2005robust} and introduced the current state of the system as decision variable to be determined as a function of the state measurement. In general, there exist other approaches apart from \cite{mayne2005robust} to guarantee recursive feasibility for linear MPC, e.g., adding additional constraints \cite{mayne2000constrained}, or using a backup strategy when the problem is infeasible \cite{hewing2018stochastic}. Guaranteeing recursive feasibility in the general case for linear systems can end up in theoretically very complicated algorithms \cite{gondhalekar2009controlled}, or it may enforce very limiting constraints that degrade the performance \cite{lofberg2012oops}.
 
In this paper, we propose a novel approach that uses MPC to generate trajectories for bipedal walking and guarantees viability of the gaits without any terminal constraint.  Our approach is based on computation of the viability kernel and projecting the measured state inside it. We ensure that, contrary to \cite{villa2017model}, the measured state is only modified if it is outside of the viability kernel.

\subsubsection{Bayesian Optimization for locomotion}
Bayesian optimization (BO) is a form of black-box and derivative-free optimization \cite{shahriari2016taking}. BO has been successfully applied to different parameter/gain tuning problems in robotics \cite{marco2016automatic,buchler2019learning}. BO is especially useful when we have a relatively low number of parameters to tune (e.g., $n \leq 20$). In other words, BO is practical when we have an \emph{efficiently searchable} control policy representation (e.g., PID \cite{buchler2019learning}, LQR \cite{marco2016automatic}, etc.) in contrast with \emph{expressive} policy representation without any prior (e.g., Neural Networks) \cite{chatzilygeroudis2019survey}.

Since humanoid robots are inherently unstable and high-dimensional systems, the application of BO to humanoid locomotion control (and in general legged robots) is limited to simplified cases such as planar bipeds \cite{calandra2014experimental,antonova2016sample,rai2018bayesian} or one-legged hoppers \cite{seyde2019locomotion}. 
\cite{calandra2014experimental} optimized eight parameters using BO for a planar biped walking. \cite{antonova2016sample} applied BO to a more complex model of a planar biped robot, where they used 16 neuromuscular policy variables to parametrize walking of a planar biped robot on uneven and sloped surfaces.
They applied a generalized version of this approach to the simulation and experiments of the biped robot ATRIAS \cite{rai2018bayesian}. 
Even for these simplified situations, both \cite{calandra2014experimental} and \cite{antonova2016sample} argued that only a very small percentage of the parameter space leads to a feasible gait, which shows the difficulty of generating feasible motions for humanoid robots using \emph{only} black-box optimization.

Recently \cite{charbonneau2018learning,yuan2019bayesian} used BO to find the parameters of a whole-body controller (inverse dynamics) for a humanoid robot, yielding robust performance for the control. Our work can be seen as complementary to these works, because we propose to use BO to find the best cost weights of the reactive planner for a given whole-body controller. In fact, we use BO to find plans that can be best tracked and can result in robustly achieving the task. 

\subsection{Proposed framework and contributions}
\begin{figure}[!t]
    \centering
    \includegraphics[clip,trim=3cm 4.6cm 10cm 4.7cm,width=.45\textwidth]{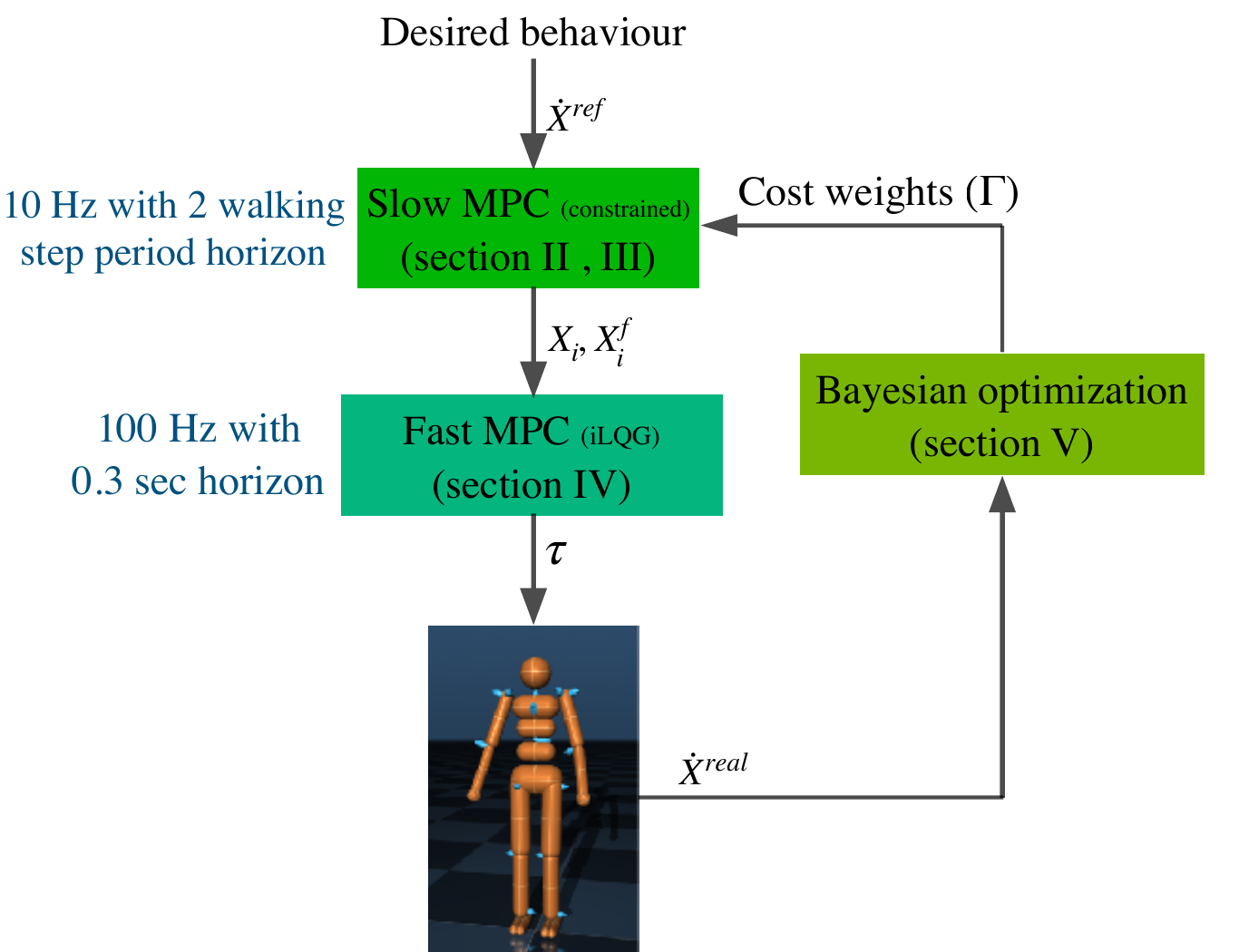}
    \caption{A high-level block diagram of the proposed approach. The first layer (slow MPC) replans the CoM and next contact location, while the second layer computes joint torques based on the full robot dynamics and the updated trajectories from the first stage. Then, BO performs a few number of simulations/experiments to find the best cost weights of slow MPC that yield a robust performance.}
    \label{fig:block_diagram}
    \vspace{-5 mm}
\end{figure}

The block diagram of our proposed control framework is shown in Fig.~\ref{fig:block_diagram}. The first layer of the MPC (slow MPC) regenerates at 10 Hz plans for the CoM and next contact location with two walking steps for the receding horizon length \cite{kajita2003biped,wieber2006trajectory,herdt2010online}. 
In the lower level, we use iterative linear-quadratic Gaussian (iLQG) \cite{tassa2012synthesis} at 100 Hz with a 0.3 s horizon (fast MPC) to generate joint torques based on the full robot dynamics and the desired trajectories from the first stage. To ensure that the first stage generates trajectories that are robust and can be tracked by the low-level controller, we use BO to find the best cost weights of slow MPC that yield a robust performance. 

In a preliminary version of this paper, we used BO to tune the cost weights of the \emph{trajectory optimization} problem \cite{yeganegi2019robust}. We showed that BO can efficiently find cost weights that lead to robust performance in the presence of different disturbances and uncertainties. 
This article extends \cite{yeganegi2019robust} in the following directions, which can be seen as the main contributions of this paper:\\
\begin{itemize}
    \item We derive boundaries of the viability kernel for LIPM with finite size foot considering the swing foot effects.
    \item We propose a viability-based projection method to modify the measured state used in the MPC to guarantee the existence of at least one solution for the OCP.
    \item We propose a two-level MPC framework (fast MPC and slow MPC, see Fig.~\ref{fig:block_diagram}) with interpretable cost terms in the slow MPC that can be learned to trade-off performance against robustness.
    \item We demonstrate the importance of feasibility of the slow MPC problem as well as generating robust trajectories, through extensive full-humanoid simulation with various realistic uncertainties. 
\end{itemize}

The rest of this paper is organized as follows: Section \ref{sec:MPC} presents the formulation that we use for slow MPC. In Section \ref{sec:initial}, we compute the boundaries of the viability kernel and propose a novel approach for projecting initial states of slow MPC inside it. Section \ref{sec:ilqg} briefly presents the fast-MPC algorithm (iLQG). In Section \ref{sec:BO}, we present the black-box optimization problem for automatically adapting the slow MPC cost weights using BO. In Section \ref{sec:res} we present extensive full-humanoid simulations to show the effectiveness of our control framework. Finally, Section \ref{sec:conclusion} summarizes the findings.

\section{Slow MPC problem}\label{sec:MPC}
In this section, we present the constrained MPC problem that we use for generating desired CoM trajectories and step locations (slow MPC in Fig.~\ref{fig:block_diagram}). We use a modified version of the MPC formulation in \cite{herdt2010online,villa2017model}:
\begin{subequations}
	\label{eq:TO}
	\begin{align}
	    \label{eq:TO,cost}
	    \underset{Z_i , X_j^f}{\text{min.}} &J \triangleq \sum\limits_{i=k}^{N+k-1}\, (\alpha \Vert \dot X_{i+1}-\dot X_{i+1}^{ref} \Vert^2  + \beta \Vert Z_{i}-Z_{i}^{ref} \Vert^2)
	    + \nonumber \\  
	    &\sum\limits_{j=1}^{m}\,(\delta \Vert X_j^f - X_j^{f,ref} \Vert^2) \quad+ 
	    \eta \Vert \xi_{k+N} - Z_{k+N}^{ref} \Vert^2 \\ \vspace{1cm}
	    \label{eq:TO,ZMP}
	    \text{s.t.}  \quad &Z_{i} \in \text{support polygon} \quad , \quad \forall i=1,...,N. \\
	    %
	    %
	    \label{eq:TO,rea}
	    \qquad &X_j^f-X_{j-1}^f \in \text{reachable area} \: , \: \forall j=1,...,m. \\
	    \label{eq:TO,swing}
	    \qquad &\frac{1}{t_{td}-t_k}(X_1^{f}-X_s^f) \in \text{swing vel. range}.
	\end{align}
\end{subequations}
where $X=[c_x,c_y]^T$ is the horizontal CoM position. $Z=[z_x,z_y]^T$ is the zero moment point (ZMP) position. $X^f=[x^f,y^f]^T$ is the footstep location ($X^f_0$ is the location of the stance foot and ${X^f_1}$ is the swing foot landing location) and $X_s^f=[x_{s}^{f},y_{s}^{f}]^T$ is the swing foot current position. $t_{td}$ is the landing time of the swing foot, while $t_k$ is the current time. $\xi=X+\frac{\dot X}{\omega_0}$ is the 2-D divergent component of motion (DCM) \cite{takenaka2009real} (which is equal to the instantaneous capture point \cite{koolen2012capturability}) and $T$ is the discretization time. Furthermore, $\Gamma=[\alpha, \beta, \delta, \eta]^T$ is the vector of cost weights, where each component is a 2D vector, e.g., $\alpha=[\alpha_x,\alpha_y]^T$. Finally, $N$ is the horizon of the MPC problem, while $m$ encodes the number of walking steps in the horizon.

In \eqref{eq:TO}, each cost term corresponds to an interpretable index. The first term in the cost \eqref{eq:TO,cost} stands for the \emph{performance}, which means that increasing $\alpha$ improves the tracking of the desired walking velocity $\dot{X}^{ref}$. The second term can be interpreted as robustness to ZMP constraints and increasing $\beta$ brings the ZMP to the center of the support polygon $Z^{ref}$ (which implicitly pushes away the ZMP from the boundaries of the support polygon). The third term prevents the footstep locations to be far from a nominal stepping with zero walking velocity $X^{f,ref}$. We can interpret the second and third terms in the cost function as \emph{constraint satisfaction robustness} terms. Finally, the last cost term guides the solutions towards being capturable at the end of the horizon, which can be seen as a \emph{stability} index: when the DCM is inside the support polygon there exists a simple controller that stabilizes the robot. Finding the best set of cost weights $\Gamma=[\alpha, \beta, \delta, \eta]^T$ is crucial, as it ensures a robust solution consistent with available uncertainties, while performance is maximized.

The constraints on the ZMP \eqref{eq:TO,ZMP} make sure that the ZMP remains inside the support foot. Constraints \eqref{eq:TO,rea} and \eqref{eq:TO,swing} ensure that the contact locations are kinematically feasible and the next step location is consistent with the current state of the swing foot and its maximum speed \cite{herdt2010online}. Using LIPM dynamics and polyhedral approximation of the friction cone, \eqref{eq:TO} can be written as a quadratic program (QP). To make sure that the QP remains always feasible, we modify the current state of the DCM based on the viability kernel bounds. We will explain the computation of the viability kernel and the proposed modification on \eqref{eq:TO} in Section \ref{sec:initial}.

\begin{remark}
In the MPC \eqref{eq:TO}, we did not take into account friction cone constraints for walking, as ZMP is the most restricting interaction constraint for walking on flat terrains \cite{wieber2016modeling} (and even rough terrain \cite{caron2017dynamic}). Adding friction cone constraints to this problem is straightforward \cite{khadiv2017pattern} and we remove them from our formulation for brevity of presentation in the viability kernel computations (Sect. \ref{subsec:viability}).
\end{remark}

\begin{remark}
Compared to \cite{villa2017model}, our MPC formulation in \eqref{eq:TO} does not take into account any uncertainty in the problem construction. We argue that finding a realistic set of uncertainties in the space of the simplified model, here LIPM, is a very challenging task. For example, it is hard (if not impossible) to map bounds on the tracking error of the swing foot or other joint space uncertainties to bounds on the CoM states. Instead, we propose to use the deterministic MPC formulation \eqref{eq:TO} in the simplified model space and add data-driven robustness to the solutions using realistic uncertainties in the full robot simulations. Furthermore, we need not make any assumption on the uncertainty structure, e.g., multiplicative or additive, with normal distribution or bounded, etc. This approach also suggests a systematic way to improve the performance using the data collected from the previous experiments. 
\end{remark}

\section{Invariance of the slow MPC}\label{sec:initial}
In this paper, we propose to use the concept of viability kernel to ensure that the CoM trajectories obtained from \eqref{eq:TO} do not diverge. A dynamical system state is said to be viable, if starting from this state there exists at least one solution that does not lead to a failed state \cite{aubin1991viability}. The set of all viable states constitutes the viability kernel, as the boundary of this set splits the state space into viable and non-viable states. The viability kernel is also called \emph{weak} forward invariant set \cite{chai2020forward}, as it does not imply that \emph{every} solution of the dynamical system starting from this set remains in this set (in this case it would be called forward invariant or positively invariant set \cite{blanchini1999set}). Note that in the case of legged robots, a more conservative concept than viability is often used, namely N-step capturability \cite{koolen2012capturability}. Viability is very similar in nature to $\infty$-step capturability, with a subtle difference that the viability kernel is a closed set while the $\infty$-step capturable region is an open set.

It is shown in \cite{wieber2008viability,wieber2016modeling} that viability of a legged robot is guaranteed, if one minimizes any derivative of the CoM motion over a \emph{long enough} horizon in the MPC cost function. However, if the measured state of the CoM is not viable (due to estimation error, measurement noise, or external disturbances), the MPC problem will diverge and tracking the resulting CoM trajectories by the whole body controller (WBC) will result in a fall.
We argue that with LIPM assumptions, which have been proven to be reasonable for walking, and given different constraints, we can compute realistic bounds for the viability kernel. As a result, we can constrain the robot state to remain inside the viability kernel of the simplified model. 

It is important to note that the viability kernel bounds based on LIPM dynamics are computed assuming zero vertical CoM acceleration and angular momentum around the CoM. Therefore, even if the measured state of the system is outside the viability kernel computed based on LIPM assumption, it is possible that the WBC brings back the state inside the viability kernel. By projecting the measured state inside the computed viability kernel we make sure that slow MPC finds a solution that does not lead to divergence of the CoM motion starting from the modified state. Then, we rely on the WBC to exploit the full control authority of the robot and bring it to a nominal safe motion.

In the remainder of this section, first we compute the viability kernel using the LIPM with constraints on the step location as well as swing foot maximum velocity. Then, we review the approach in \cite{villa2017model} and propose an alternative approach to modify the OCP \eqref{eq:TO} to ensure weak invariance of the motion while remaining always feasible. Finally we comment on the differences and implications of each approach.

\begin{remark}
    In our MPC formulation, we considered two walking steps as the horizon. However, as shown in \cite{khadiv2020walking}, one step horizon could be enough, provided that a proper terminal constraint is considered that accounts for the viability of the gait. Finding this time invariant terminal constraint for the LIPM is practical, when the contact switch is the terminal point in the horizon. In that case, we end up having a shrinking horizon MPC formulation. To resort to the standard fixed horizon MPC formulation, the common practice traditionally was to just consider two walking steps in the horizon as walking is a two-step periodic gait. In this approach, the hope is that minimizing any derivative of the CoM velocity is enough to ensure gait invariance \cite{wieber2008viability}. However, it is shown in \cite{scianca2020mpc} that a terminal constraint is essential to guarantee the invariance of the gait. Finding this terminal constraint amounts to making some assumptions on the gait after the terminal time (the tail of the MPC). Adding a terminal constraint also makes it essential to find a way to guarantee recursive feasibility. In this paper, we use two steps in the horizon without any terminal constraint and leverage the approach in \cite{khadiv2020walking} for computing the viability kernel to provide the MPC formulation in (1) with viability guarantees.
\end{remark}

\subsection{Viability kernel}\label{subsec:viability}
In this section we follow procedures similar to \cite{khadiv2020walking} to compute the bounds of the viability kernel for the case of an LIPM with finite-size foot (rectangular shape) and box constraints for the step location. We could also compute this bound analytically (or through a convex optimization problem) with the same procedure for other convex constraints on foot step location and other convex foot shapes.
Compared to \cite{khadiv2020walking,koolen2012capturability} that did not consider the current state of the swing foot, here we take into account constraint \eqref{eq:TO,swing} to compute the viability kernel. In this way we can make sure that the viability kernel bounds are consistent with the reachability constraint of the current state of the swing foot. 

Starting from \eqref{eq:TO,swing}, we can compute the reachable area for the landing of the swing foot in sagittal direction
\begin{align}\label{eq:sagital_landing}
		\abs{x_1^f-x_s^f}\leq \overline{v}_x (t_{td}-t)
\end{align}
where $t_{td}$ is the touch down time and $\overline{v}_x$ is the maximum average foot velocity. Note that $\overline{v}_x$ is a simplified upper-bound on the average velocity of the swing foot proposed in \cite{herdt2010online} as a proxy constraint that is used in \eqref{eq:TO,swing}. Note also that, we use underline and overline to show minimum and maximum of a variable all over the paper. Using (\ref{eq:sagital_landing}), we can compute the maximum and minimum reachable locations for the swing foot landing, given the maximum swing foot velocity
\begin{subequations}
\label{eq_app:swing_max_x}
\begin{align}
    \overline{x}_1^f = x_s^f + \overline{v}_x (t_{td}-t)
    \\
    \underline{x}_1^f = x_s^f - \overline{v}_x (t_{td}-t)
\end{align}
\end{subequations}
Taking into account the kinematic reachability constraint for the step length, we compute the maximum and minimum step length for the current step as
\begin{subequations}
\label{eq_app:foot_max_x}
\begin{align}
    \overline{x}_1^{f,rea} = \; &(x^f_1 - x^f_0)_{max} 
    = \; \text{min } (\overline{x}_1^f - x^f_0,\overline{L}) \\
	\underline{x}_1^{f,rea} = \; &(x^f_1 - x^f_0)_{min} 
    = \; \text{max } (\underline{x}_1^f - x^f_0, -\overline{L})
\end{align}
\end{subequations}
Where $\overline{L}$ is the maximum step length and the superscript $rea$ stands for reachable.

Using the results in \cite{khadiv2020walking}, we can write down the evolution of the DCM offset $b$, which is the distance between the center of the stance foot and the DCM. When the DCM is in front of the stance foot (i.e. $b_{t,x} \geq \frac{L_f}{2}$, with $L_f$ being the foot length), the \emph{best} ZMP location to slow down the DCM divergence is the tip of the foot. This results in the following DCM evolution (see Appendix A for details):
\begin{subequations}
\label{eq_app:DCM-current}
\begin{align}
    b_{T,x} - \frac{L_f}{2} + x^f_1= \left(b_{t,x} -\frac{L_f}{2}\right) e^{\omega_0(T_s-t)}+ x^f_0
\end{align}
where $b_t$ and $b_T$ are the DCM offset at the beginning of the current time and next step, $T_s$ is the single support period.
When the DCM is instead behind the stance foot, i.e. $b_{t,x} \leq -\frac{L_f}{2}$, we have: 
\begin{align}
    b_{T,x} + \frac{L_f}{2} + x^f_1= \left(b_{t,x} +\frac{L_f}{2}\right) e^{\omega_0(T_s-t)}+ x^f_0
\end{align}
\end{subequations}
From these equations we can compute $\overline{b}_{t,x}$, the boundary of the viability kernel in sagittal direction as a function of the maximum/minimum step length $(x^f_1 - x^f_0)_{max/min}$ and the maximum DCM offset at the next step $\overline{b}_{T,x}$:
\begin{subequations}
\label{eq_app:DCM-offset-current}
\begin{align}
    \overline{b}_{t,x} =\frac{\overline{b}_{T,x}-L_f/2 + (x^f_1 - x^f_0)_{max}}{e^{\omega_0(T_s-t)}}+\frac{L_f}{2}
    \\
    \underline{b}_{t,x} =\frac{-\overline{b}_{T,x}+L_f/2 + (x^f_1 - x^f_0)_{min}}{e^{\omega_0(T_s-t)}}-\frac{L_f}{2} 
\end{align}
\end{subequations}
Since $\overline{b}_{T,x}$ refers to the beginning of the (next) step, we can reasonably assume that it is independent of the swing foot state. Therefore, we know from \cite{khadiv2020walking} that it can be computed as:
\begin{align}
    \overline{b}_{T,x} = \frac{\overline{L}}{e^{\omega_0 T_s}-1}+\frac{L_f}{2}\nonumber
\end{align}
\begin{center}
\begin{figure*}[!tbp] 
\centering
\setlength{\belowcaptionskip}{0mm} 
\includegraphics[clip,trim=1.9cm 42.3cm 1.7cm 5cm,width = \textwidth]{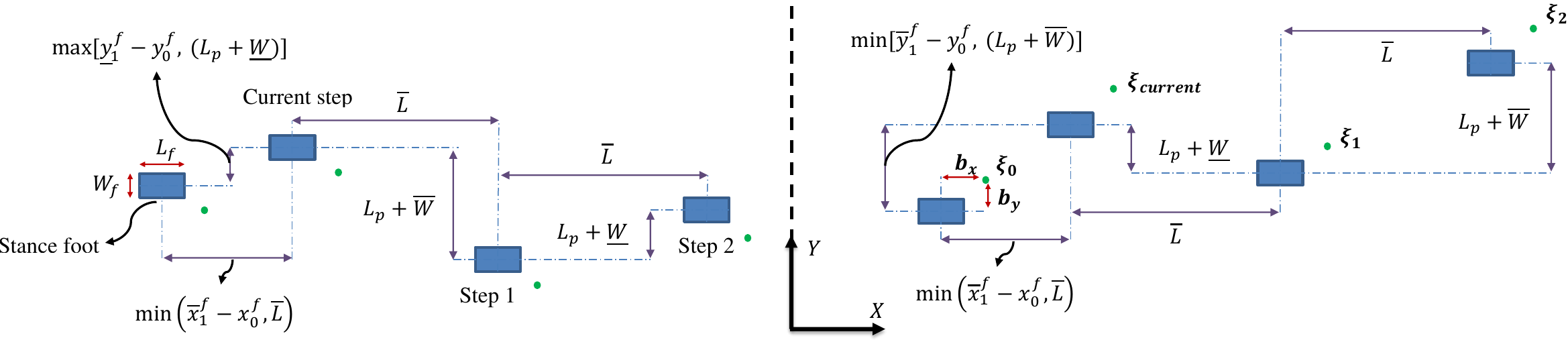}
\caption{A schematic view of the walking pattern showing the footprints, the DCM, and the DCM offset. The right foot is in stance. (left) The DCM is on the right of the stance foot. (right) The DCM is on the left of the stance foot.}
\vspace{-0.5em}
\label{fig:DCM}
\end{figure*}
\end{center}

Substituting this equation and \eqref{eq_app:foot_max_x} into \eqref{eq_app:DCM-offset-current} yields
\begin{subequations}
\label{eq:viability-sagittal}
\begin{align}
    \overline{b}_{t,x} = \left ( \frac{\overline{L}}{e^{\omega_0 T_s}-1}+\overline{x}_1^{f,rea} \right )\, e^{-\omega_0(T_s-t)}+\frac{L_{f}}{2}
    \\
    \underline{b}_{t,x} = \left (\frac{-\overline{L}}{e^{\omega_0 T_s}-1}+\underline{x}_1^{f,rea} \right )\, e^{-\omega_0(T_s-t)}-\frac{L_{f}}{2}
\end{align}
\end{subequations}
where $\overline{L}$ is the maximum step length, $\overline{x}_1^{f,rea}$ and $\underline{x}_1^{f,rea}$ are the maximum and minimum reachable locations for the swing foot in sagittal direction computed in \eqref{eq_app:foot_max_x}. Finally, $\overline{b}_{t,x}$ is the maximum DCM offset at time $t$ in sagittal direction. To apply the viability constraint in sagittal direction, the DCM offset ($b_{t,x}$) must lie between the bounds of \eqref{eq:viability-sagittal}.

Equation \eqref{eq:viability-sagittal} is a generalization of the viability kernel bounds computed in \cite{khadiv2020walking}, that takes into account the maximum swing foot velocity, as well as the foot length effect. We can verify that the bounds of \cite{khadiv2020walking} are a special case of~\eqref{eq:viability-sagittal} by setting $T_s-t=T_s$, $L_{f}=0$ and $\overline{x}_1^{f,rea}=\overline{L}$: 
\begin{align}
    \overline{b}_{T,x} = \frac{\overline{L}}{e^{\omega_0 T_s}-1} \nonumber
\end{align}
In lateral direction, we have asymmetric reachability constraint sets because of self-collision constraints. Hence, we consider two cases for computing the viability kernel, i.e. \emph{inward direction} for the case where swing foot adaptation is prone to self-collision, and \emph{outward direction} where the swing foot is only limited by kinematic reachability constraint. The viability kernel bounds for the lateral direction are (see the Appendix B for derivation details)
\begin{subequations}
\label{eq:viability-sagittal-foot-y}
    \begin{align}
    \overline{b}_{t,y}^{in} =& \bigg[ y_1^{f,rea,in}
    (-1)^n\frac{L_p}{1+e^{\omega_0 T_s}} + \nonumber\\
    (-1)^n&\frac{\overline{W}-\underline{W}e^{\omega_0 T_s}}{1-e^{2\omega_0 T_s}} \bigg] e^{-\omega_0 (T_s-t)}
    (-1)^{n-1}\frac{W_f}{2}\\
    \overline{b}_{t,y}^{out} =& \bigg[y_1^{f,rea,out}
    (-1)^n\frac{L_p}{1+e^{\omega_0 T_s}} + \nonumber\\
    (-1)^n&\frac{\underline{W}-\overline{W}e^{\omega_0 T_s}}{1-e^{2\omega_0 T_s}}\bigg]e^{-\omega_0 (T_s-t)}
    (-1)^n\frac{W_f}{2}
    \end{align}
\end{subequations}
where $y_1^{f,rea,in}$ is computed using \eqref{eq_app:foot_max_right_outward_y} or \eqref{eq_app:foot_max_left_outward_y} and $y_1^{f,rea,out}$ is computed using \eqref{eq_app:foot_max_right_inward_y} or \eqref{eq_app:foot_max_left_inward_y} based on which foot is stance, $W_f$ is the foot width. $\underline{W}$ and $\overline{W}$ are the minimum and maximum step width with respect to the nominal step width which is pelvis width $L_p$ (note that $\underline{W}$ and $\overline{W}$ could also be negative \cite{khadiv2020walking}); $n=1$ when the right foot is stance, and $n=2$ when the left foot is stance. Again, setting $T_s-t=T_s$, $W_f=0$, $y_1^{f,rea,in} = (L_p+\overline{W})$, and  $y_1^{f,rea,out} = (L_p+\underline{W})$ and assuming the right foot is stance, we obtain the same result as in \cite{khadiv2020walking}
\begin{subequations}
    \begin{align}
    \overline{b}_{t,y}^{in} &= \frac{L_p}{e^{\omega_0 T_s}+1} +\frac{\underline{W}-\overline{W}e^{\omega_0 T_s}}{1-e^{2\omega_0 T_s}}\nonumber\\
    \overline{b}_{t,y}^{out} &= \frac{L_p}{e^{\omega_0 T_s}+1} +\frac{\overline{W}-\underline{W}e^{\omega_0 T_s}}{1-e^{2\omega_0 T_s}}\nonumber
    \end{align}
\end{subequations}

\subsection{Approach I: Initial condition as decision variable}\label{subsec:approach1}
The first approach we consider to ensure a non-divergent motion of the CoM in  \eqref{eq:TO} is similar in spirit to the one proposed in \cite{mayne2005robust} and used in \cite{villa2017model} for bipedal walking. In this approach, we add a conservative terminal constraint to \eqref{eq:TO}, i.e. capturability. Then, in order to make sure that the OCP in \eqref{eq:TO} remains always feasible, we consider the initial states of the CoM as decision variable.
The modified OCP is
\begin{align}\label{eq:TO_modified}
    \underset{X_0, \dot{X_0}, Z_i , X_j^f}{\text{minimize}} \quad & J\\  
    \text{s.t.} \quad  &\eqref{eq:TO,ZMP},\eqref{eq:TO,rea},\eqref{eq:TO,swing}.\nonumber \\
    & \xi_{k+N} \in \text{support polygon} \nonumber
\end{align}
In this formulation, the initial state is allowed to be changed such that the OCP remains always feasible. The initial state can be arbitrarily selected by the program such that the cost is minimized and in this way the feasibility of the program is guaranteed by construction.

As we will show later, since the initial condition can be selected arbitrarily by the optimizer, using \eqref{eq:TO_modified} can result in a discontinuous trajectory of the CoM. One can think of adding a cost term to reward initial states that are close to the measured state; however, in practice we observed that we would need a very high weight for this term to have a smooth CoM trajectory. This would be problematic for the tuning of our (interpretable) cost terms to be weighted automatically using BO. To circumvent this, we propose an alternative approach that does not suffer from this problem.

\subsection{Approach II: Projection of measured state inside viability kernel}\label{subsec:approach2}

In the second approach, we propose a new way to adapt the measured state to guarantee viability while remaining always feasible. We construct the following QP to project the measured (estimated) CoM states $X_{0}^{mea}$, $\dot X_{0}^{mea}$ inside the viability kernel before passing it to \eqref{eq:TO}
\begin{subequations}
	\label{eq:state_projection}
	\begin{align}
	    \label{eq::state_projection,cost}
	    \underset{X_0, \dot{X_0}}{\text{minimize}} \quad & \Vert X_{0}-X_{0}^{mea} \Vert^2 + w \Vert \dot X_{0}-\dot X_{0}^{mea} \Vert^2\\  
	    \label{eq::state_projection,viability}
	    \text{s.t.} \quad  & \xi_0=X_0+\frac{1}{\omega_0} \dot{X}_0 \in \text{viability kernel}.
	\end{align}
\end{subequations}
where $w$ is a constant weight. In this way we can guarantee existence of at least one solution for \eqref{eq:TO} that does not lead to a divergence of the CoM motion, starting from initial state $X_{0}$, $\dot X_{0}$ computed by \eqref{eq:state_projection}. This QP will simply output the measured state as long as it is viable, otherwise it will project the measured state inside the viability kernel. This is a desired behaviour compared to the previous approach in Section \ref{subsec:approach1} where the initial condition can be arbitrarily selected by the OCP to minimize the cost function. In Section \ref{sec:res}, we will compare the performance of each approach in both LIPM simulation and full-humanoid simulation.

\begin{remark}
We proposed to use the viability kernel bounds to map the measured state of the system to the viability kernel. One might wonder why we did not also use the viability kernel as a terminal constraint set of the MPC (1). The answer is since the bounds we computed for viability are functions of the swing foot state and we do not know the state of the swing foot at the end of the horizon (and the optimal location of the step after that), we cannot know the viability bounds at the end of the horizon. There are three ways around this problem; 1) Assuming the state of the swing foot at the end of the horizon is equal to its current state (as the horizon of our MPC is exactly two walking steps). However, it might be the case that the robot is disturbed (or the commanded walking velocity is changed), hence the step locations are not periodic anymore, leading the state of the swing foot to be different in two steps. 2) Changing the MPC horizon at each time such that the start of the second step in the horizon is always the end of the horizon. In this case there would be no state of the swing foot of the last step in the horizon, and we could use the viability bounds computed without considering the swing foot. However, we would need to change the matrix sizes and the problem structure at every loop and solve a shrinking horizon MPC rather than a well-understood fixed horizon MPC. 3) Adding viability constraints at the switching times which are independent of the swing foot states. This way we would only need to change inequality constraint matrices and solve a fixed horizon MPC. But again in this case we are making the terminal state free, and in our experiments we could not find a noticeable difference between the result of this approach and the one we propose in the paper, i.e. to just project the current state of the CoM into the viability kernel, while having a capturability terminal cost, whose weight is traded-off against other terms using BO. As a result, we decided not to use any of the above three options.
\end{remark}{}

\section{Fast MPC problem}\label{sec:ilqg}

We use iLQG to track the CoM and feet trajectories obtained using slow MPC. The controller maps the desired CoM and feet trajectories from the first stage to joint torques, while penalizing the full-body constraints~\cite{tassa2012synthesis,tassa2014control}. 
Typically, an inverse dynamics/kinematics controller is used to track the trajectories \cite{khadiv2016step,ponton2018time,carpentier2018multicontact}, but instead we use iLQG in a closed-loop MPC fashion, i.e. we solve the OCP starting from the measured state of the robot. The reasons behind this choice are 1) to enable the whole-body controller to find a consensus between foot trajectory and CoM trajectory, being able to generate realistic angular momentum trajectories, 2) to make the generated control commands less aggressive, 3) thanks to recent advances in DDP-like algorithms based on analytical derivatives of rigid body dynamics \cite{mastalli2019crocoddyl}, solving the whole-body MPC problem in real-time is becoming computationally feasible.

In our problem, the cost function consists of several error terms representing the following desired tasks. The values in parentheses show the range of the cost weights for each task we used in all our simulations:
\begin{itemize}
	\item Smooth-abs~\cite{tassa2012synthesis} of the tracking error of reference feet and CoM trajectories $(100, 500)$.
	\item Square of two-norm of the tracking error of reference feet and CoM velocities $(50, 500)$.
	\item Square of two-norm of joint torques $(100)$, joint velocities $(0.001, 10)$, angular velocity of the pelvis $(10,100)$, and linear velocity of the torso in vertical direction $(300)$.
	\item Square of two-norm of deviation between the global orientations and the orientations of the torso $(100)$, pelvis $(100)$, and feet $(10, 100)$.
	\item Square of two-norm of the deviation of the global height of the torso from the fixed value used for the LIPM $(300)$.
	\item Square of two-norm of the deviation between the joint configuration and the initial posture $(0,100)$. 
\end{itemize} 

\section{Finding Optimal cost weights of slow MPC}\label{sec:BO}

Based on the slow MPC formulation in \eqref{eq:TO}, each cost term stands for an interpretable index, namely performance (first term), constraint satisfaction robustness (second and third terms), and stability (last term). As a result, in the presence of different uncertainties, one can choose different weights $\Gamma=[\alpha, \beta, \delta, \eta]^T$ that can result in a robust performance. For instance, let us consider the second cost term in \eqref{eq:TO}. If we set $\beta$ to zero, the ZMP could likely reach the boundaries of the support polygon, as normally the ZMP constraint \eqref{eq:TO,ZMP} is the most restrictive constraint in \eqref{eq:TO}. If we start increasing $\beta$, we favor solutions of the program in \eqref{eq:TO} that have a ZMP close to the center of support polygon. However, increasing this weight too much not only prevents the robot from achieving motions with high walking velocities, but it may also activate other constraints or even jeopardise stability. Hence, with a proper choice of cost weights, we can make sure that the optimal solution is the one that yields a robust performance.

\subsection{Problem formulation}

We are interested in solving the following optimization problem (see Fig.~\ref{fig:block_diagram}):
\begin{align}
\label{eq:opt_final}
\underset{\Gamma}{\text{minimize}\nonumber} & \, J_{BO}(\Gamma)\triangleq \sum\limits_{i=1}^{N}\, \Vert \dot X^{\text{real}}_{i}(\Gamma) -\dot X_{i}^{des} \Vert^2 
\\
\text{s.t.} \quad & \dot X^{\text{real}}_{i}(\Gamma)\, \text{is the output of simulation in Fig.~\ref{fig:block_diagram},}
\end{align}
where $\dot X^{\text{real}}_{i}(\Gamma)$ is the CoM velocity obtained from the simulation of the robot full body with different (unknown) disturbances (or real experiment). In fact, here we are considering the whole control procedure as a black box, where the cost weights of the slow MPC are the parameters of the policy that are decided by the BO.

If the robot falls down during one episode (the difference between the desired and actual heights of the CoM exceeds a constant threshold), we terminate that episode and return a high penalty as the cost of the episode. This penalty, which is chosen to be larger than the worst tracking performance without falling down, is defined so that later falls receive smaller penalty. 

\subsection{Bayesian optimization}
In order to solve \eqref{eq:opt_final}, we resort to BO, which has shown to be very efficient for problems with a low number of parameters to learn \cite{calandra2014bayesian,marco2016automatic,rai2018bayesian,buchler2019learning}. Note that since we do not make any assumption in terms of the uncertainties and disturbances (we do black-box optimization), our approach is not limited to uncertainties with a given shape or probability distribution, which is the case for RMPC and SMPC problems \cite{villa2017model,gazar2020stochastic}. On the downside, contrary to \cite{villa2017model,gazar2020stochastic}, we need to carry out a few simulation experiments to achieve the robustness and any change in the uncertainty set would need a new set of simulation experiments.

BO is one of the most efficient algorithms for active learning of policy parameters. In a nutshell, BO builds a \emph{surrogate model} of the cost function (in our case \eqref{eq:opt_final}) typically using Gaussian processes (GP). Then, it optimizes an \emph{acquisition function} which is based on the surrogate model to find the next set of parameters, given the history of the experiments until now. The acquisition function tries to find a trade-off between exploration (high-variance) and exploitation (high-value). 

To solve the BO problem in this paper we use scikit-optimize\footnote{\url{https://github.com/scikit-optimize/scikit-optimize}} and employ $gp-hedge$ as acquisition function, which is a probabilistic combination of lower confidence bound, expected improvement and probability of improvement \cite{hoffman2011portfolio}. More details on the BO technique that we used can be found in \cite{yeganegi2019robust}.

\section{Results and Discussion}\label{sec:res}

In this section we present simulations of a full humanoid robot to showcase the effectiveness of our framework. We used MuJoCo \cite{todorov2012mujoco} for all our full-body simulations on a laptop with 3.6 GHz Intel i7 processor and 16Gb of RAM. The considered humanoid robot is 1.37 m tall, it weighs 41 kg and has 27 DoFs. Abdomen, shoulder, and ankle joints are 2-DoF, while elbows, knees, and pelvis are 1-DoF, and hips are 3-DoF joints. Table \ref{tab:LIPM-properties} summarizes the physical properties of the LIPM approximation of the robot and the gait parameters used in walking simulations.

\begin{table}[!tbp]
\centering
\caption{Physical properties of LIPM and gait parameters}
\begin{tabular}{@{}ccc@{}}
\toprule
Parameter & Description             & Value     \\ \midrule
$h$                             & LIPM height                     & 0.8 $(m)$                   \\
$L_f$                             & Foot length                     & 0.2 $(m)$                   \\
$W_f$                             & Foot width                      & 0.1 $(m)$                   \\
$L_p$                             & Pelvis length                     & 0.2 $(m)$                   \\
$\overline{L}$                             & Maximum step length             & 0.6 $(m)$                   \\
$L_p + \overline{W}$                             & Maximum step width             & 0.4 $(m)$                   \\
$L_p + \underline{W}$                            & Minimum step width              & 0.12 $(m)$                  \\
$T_{ss}$                             & Single support duration         & 0.5 $(s)$                 \\
$T_{ds}$                             & Double support duration         & 0.1 $(s)$                 \\
$dT$                             & Time step                       & 0.1 $(s)$                 \\ \bottomrule
\end{tabular}
\label{tab:LIPM-properties}
\end{table}

As shown in Fig.~\ref{fig:block_diagram}, we regenerate the CoM and feet trajectories every 0.1 s using the LIPM with a two-walking-step horizon (slow MPC), and the iLQG every 0.01 s for a 0.3 s horizon (fast MPC), both in closed-loop. Note that the discretization times are the same as the replanning, i.e. 0.1 s for slow MPC and 0.01 s for fast MPC. Simulations are run at 1 KHz. Note that although we re-compute the iLQG policy every 10 ms and the feedforward and feedback terms are fixed during this period, we update the measured/estimated state of the system every 1 ms in the control law
\begin{align}
\label{eq:ilqg-update-torques}
&u_k = u_i^* + K_i (x_k - x_k^*)
\end{align} 
where $x_k$ and $u_k$ are the state and control input at time step $k$ (updated every $0.001$ s), $K$ is the feedback gain matrix, and $(x^*,u^*)$ is the optimal state-control trajectory obtained from iLQG. 

In the rest of this section, we present three different sets of simulations to illustrate the capabilities of the proposed framework. In Subsection \ref{sec:res,viability}, we show the effectiveness of the projection stage to ensure feasibility of the MPC problem. In Subsection \ref{sec:res,uncertainty}, we investigate the effects of different uncertainties, i.e. computational delay and model uncertainties on the performance of the control framework. Finally, in Subsection \ref{sec:res,BO}, we show the effectiveness of BO in finding the best cost weights of the MPC through a few episodes in simulation.

\subsection{Viability-based projection}\label{sec:res,viability}

In this subsection, we investigate the effectiveness of the viability projection described in Section \ref{sec:initial}. To do that, first we compare our projection method, namely Approach II presented in Section \ref{subsec:approach2} to the Approach I presented in Section \ref{subsec:approach1} which is similar to the one proposed in \cite{mayne2005robust} and used in \cite{villa2017model} for locomotion. We show systematically that Approach II outperforms Approach I, and use that in the rest of the simulations. Then, we compare the performance of our projection method to the traditional MPC problem without projection. 
For all the simulations in this section, we used a combination of stepping in place, walking with a desired forward velocity, and going back to zero velocity stepping in place. 

\begin{figure}[ht]
\centering 
\includegraphics[width = \columnwidth]{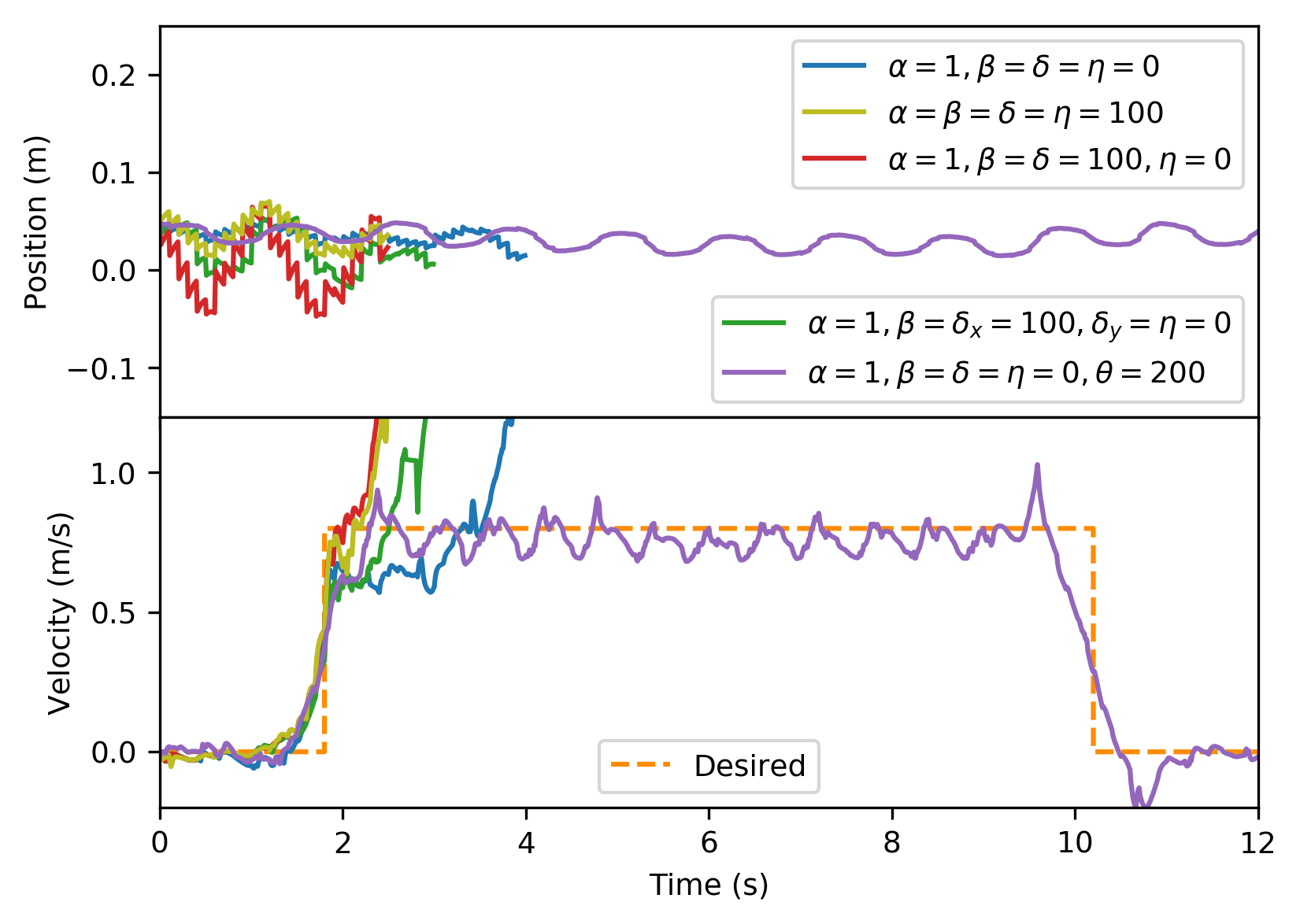}
\vspace{-5mm}
\caption{
(top) CoM position and (bottom) velocity tracking performance with projection based on \emph{Approach I}.
}
\label{fig:s1c31-Approach_I_without_cost_term}
\end{figure}

\begin{figure}[!tbp]
\centering 
\includegraphics[width = \columnwidth]{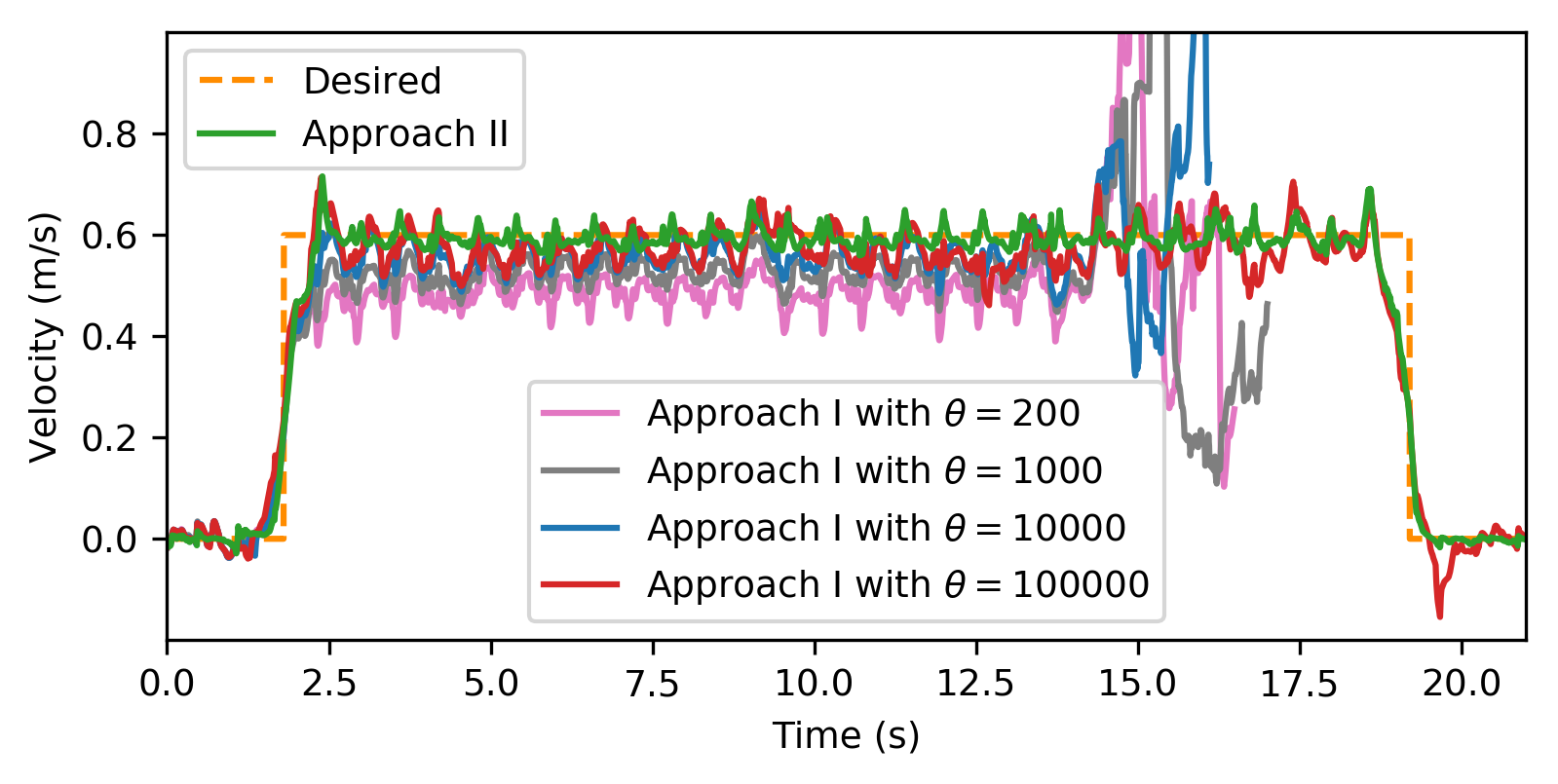}
\vspace{-5mm}
\caption{ Comparison between two approaches for the projection. 
}
\label{fig:s1c31-Approach_I_with_cost_term}
\end{figure}

\begin{center}
\begin{figure*}[!tbp]
\begin{subfigure}{\textwidth}
\centering 
\includegraphics[width =\textwidth]{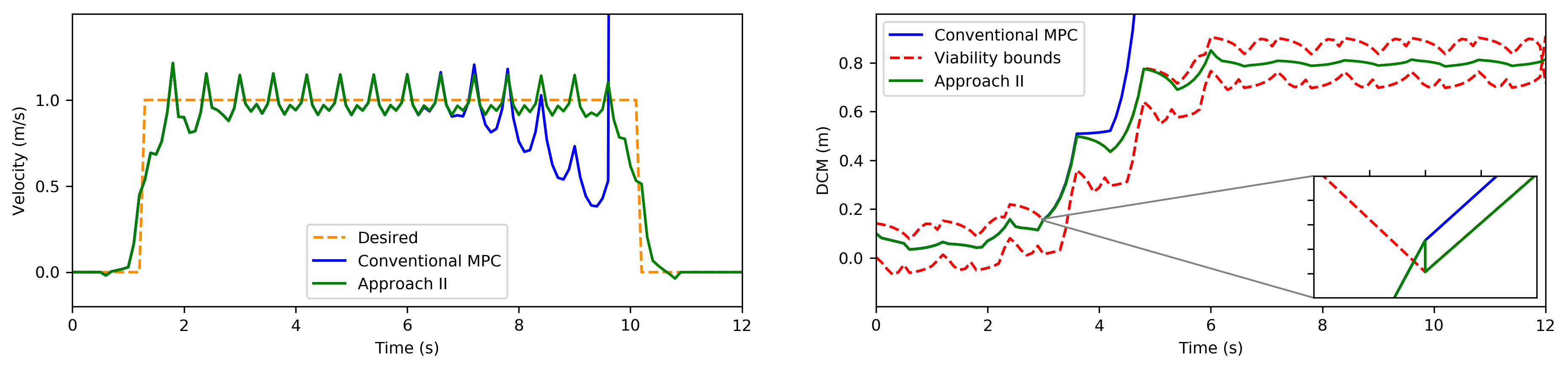}
\vspace{-5mm}
\caption{Simplified model (LIPM) simulation with one external push.}
\label{fig:s1c1_1}
\vspace{2mm}
\end{subfigure}

\begin{subfigure}{\textwidth}
\centering 
\includegraphics[width =\textwidth]{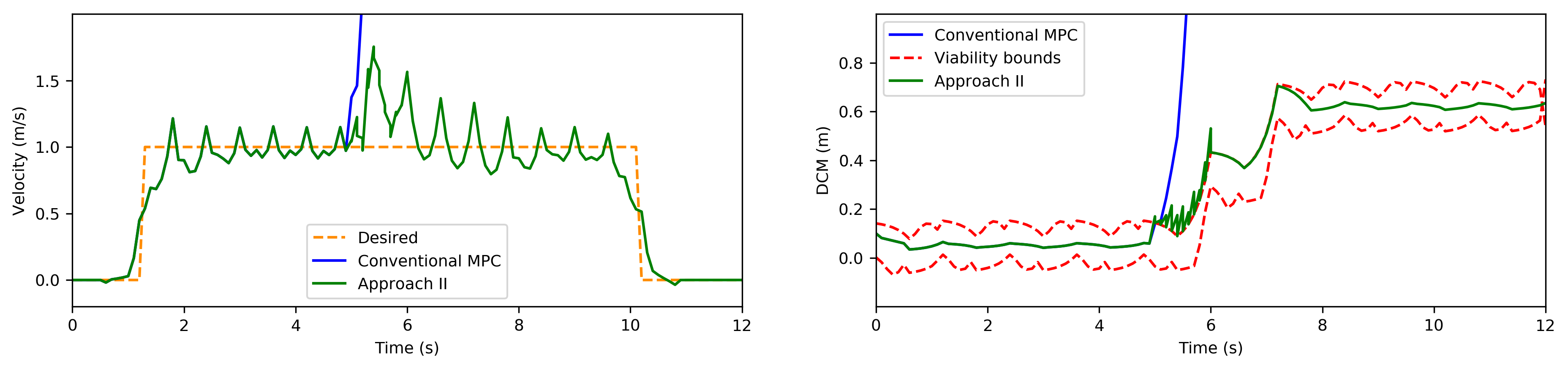}
\vspace{-5mm}
\caption{Simplified model (LIPM) simulation with several external pushes.}
\label{fig:s1c1_2}
\vspace{2mm}
\end{subfigure}

\begin{subfigure}{\textwidth}
\centering 
\includegraphics[width =\textwidth]{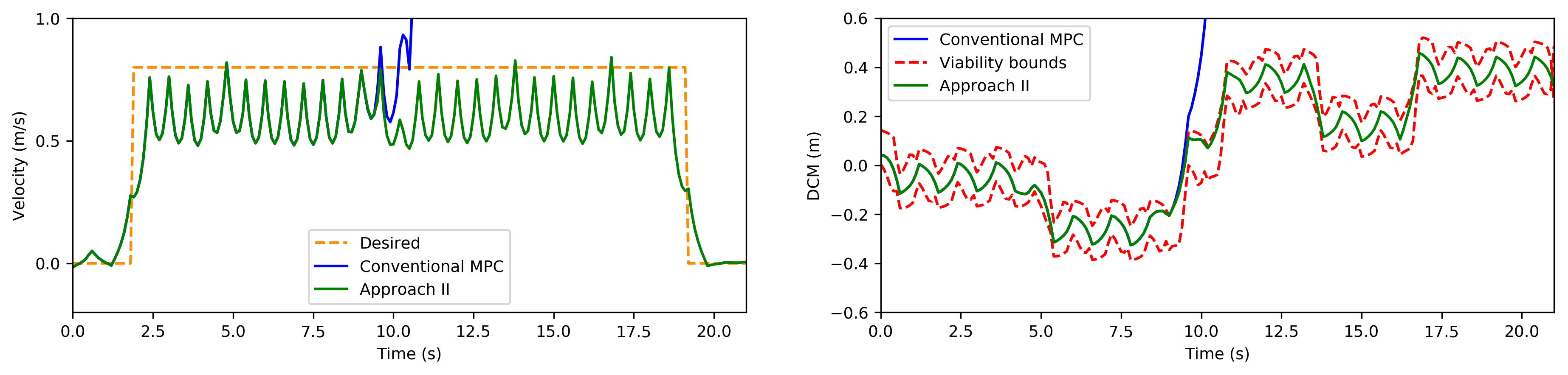}
\vspace{-5mm}
\caption{Full-body robot simulation with external push.}
\label{fig:s1c2}
\end{subfigure}
\vspace{2mm}
\caption{Comparison between MPC with the projection based on \emph{Approach II} and without projection, (a) LIPM simulation with one lateral push (b) several lateral pushes, (c) Full body simulation.}
\label{fig:s1c1&2-velocity-DCM}
\end{figure*}
\end{center}

To compare the performance of Approach I and Approach II, first we used Approach I in the full-body robot simulation with several sets of cost weights ($\alpha, \beta, \delta, \eta$). In this approach, since the initial state is free to be chosen by the solver to minimize the cost function, we ended up with discontinuous trajectories as shown in Fig.~\ref{fig:s1c31-Approach_I_without_cost_term}, which resulted in instability when applied to the full-body simulation. We tested this projection with different weights of the cost function and observed the same behaviour as shown in Fig.~\ref{fig:s1c31-Approach_I_without_cost_term}. To solve this problem we defined a new cost term in \eqref{eq:TO_modified} with the cost weight of $\theta$, which incentivizes the initial state towards the measured state. When the cost weights are $\alpha_x=\alpha_y=1,\beta_x=\beta_y=\delta_x=\delta_y=\eta_x=\eta_y=0$, our tests showed that it is sufficient to set $\theta$ to a value greater than 200 to make the CoM trajectories smooth enough to be tracked in the full-body simulation and achieve walking.

There are three main drawbacks for this approach. First, the new cost term has an explicit effect on MPC performance. Therefore, its cost weight has to be re-tuned carefully to generate a smooth trajectory when the other cost weights change. Second, it is desirable that the MPC starts from the measured state, as long as the current state is viable. Considering the initial state as decision variable and adding a cost to enforce this may not necessarily provide this in all situations. Third, the cost weights are our design parameters for the BO problem (which are index of performance or robustness). Hence, it is preferable to exclude this extra term from the cost function. Despite these caveats, we have conducted BO for Approach I with $\theta$ as a new design variable and have observed that Approach II outperforms it in the presence of different realistic uncertainties (see \ref{sec:res,BO}).

To investigate this problem more clearly, we implement Approach I in a full-body simulation with four lateral pushes $F_d=-35$ N, $F_d=45$ N, $F_d=-60$ N, and $F_d=70$ N uniformly to the robot at $t=4.2$ s, $t=8.4$ s, $t=13.2$ s, and $t=16.2$ s, during $\Delta t = 0.2 $ s. The cost weights that we considered are $\alpha_x=\alpha_y=1000,\beta_x=\beta_y=\delta_x=\delta_y=\eta_x=\eta_y=0$. Among different cost weights we tested and shown in Fig.~\ref{fig:s1c31-Approach_I_with_cost_term}, only with $\theta = 100000$ the robot can recover from the pushes and avoid falling. Thus, finding a proper value for $\theta$ is challenging and reasonable values change when changing the other cost weights. However, as shown in Fig.~\ref{fig:s1c31-Approach_I_with_cost_term}, Approach II generates trajectories that are tracked well in the simulation, without adding any cost term and adding complexity to the problem. 

To demonstrate how Approach II works, we first perform an LIPM simulation. In this simulation, we set the cost weights of \eqref{eq:TO} as $\alpha_x=\alpha_y=1,\beta_x=\beta_y=\delta_x=\delta_y=\eta_x=\eta_y=0$. We disturb the state of the LIPM in $y$ direction at $t=2.0$ s and $t=3.0$ s with
$
\Delta c_y=0.01$ m, $\Delta \dot c_y=0.05$ m/s
and
$
\Delta c_y=0.02$ m, $\Delta \dot c_y=0.1$ m/s
, and compare the output of our MPC with viability projection (Approach II) to the one without projection (named conventional MPC). Note that when we use Approach II, we manually set the measured state (the LIPM simulation outputs) the same as the projected values, once the states are projected inside the viability kernel using \emph{Approach II} (we do this only for the LIPM simulations, i.e. Fig.~\ref{fig:s1c1_1} and Fig.~\ref{fig:s1c1_2}). This means that we assume the whole-body controller would be able to track the desired CoM trajectory.
In Fig.~\ref{fig:s1c1_1} (right) after the second disturbance at $t = 3.0$ s, the DCM in lateral direction exits the viability kernel slightly. With the conventional approach and no projection stage, the DCM diverges towards infinity (Fig.~\ref{fig:s1c1_1}, left). Instead, \emph{Approach II} projects the state once at $t = 3.0$ s only in the lateral direction when the disturbance is applied. As a result, the DCM remains close to the upper limit for a while and goes back inside the boundaries again at around $t = 5.0$ s (Fig.~\ref{fig:s1c1_1} right and Fig.~\ref{fig:s1c1_1} left). 

\begin{center}
\begin{figure*}[!tbp] 
\includegraphics[width = .95\textwidth]{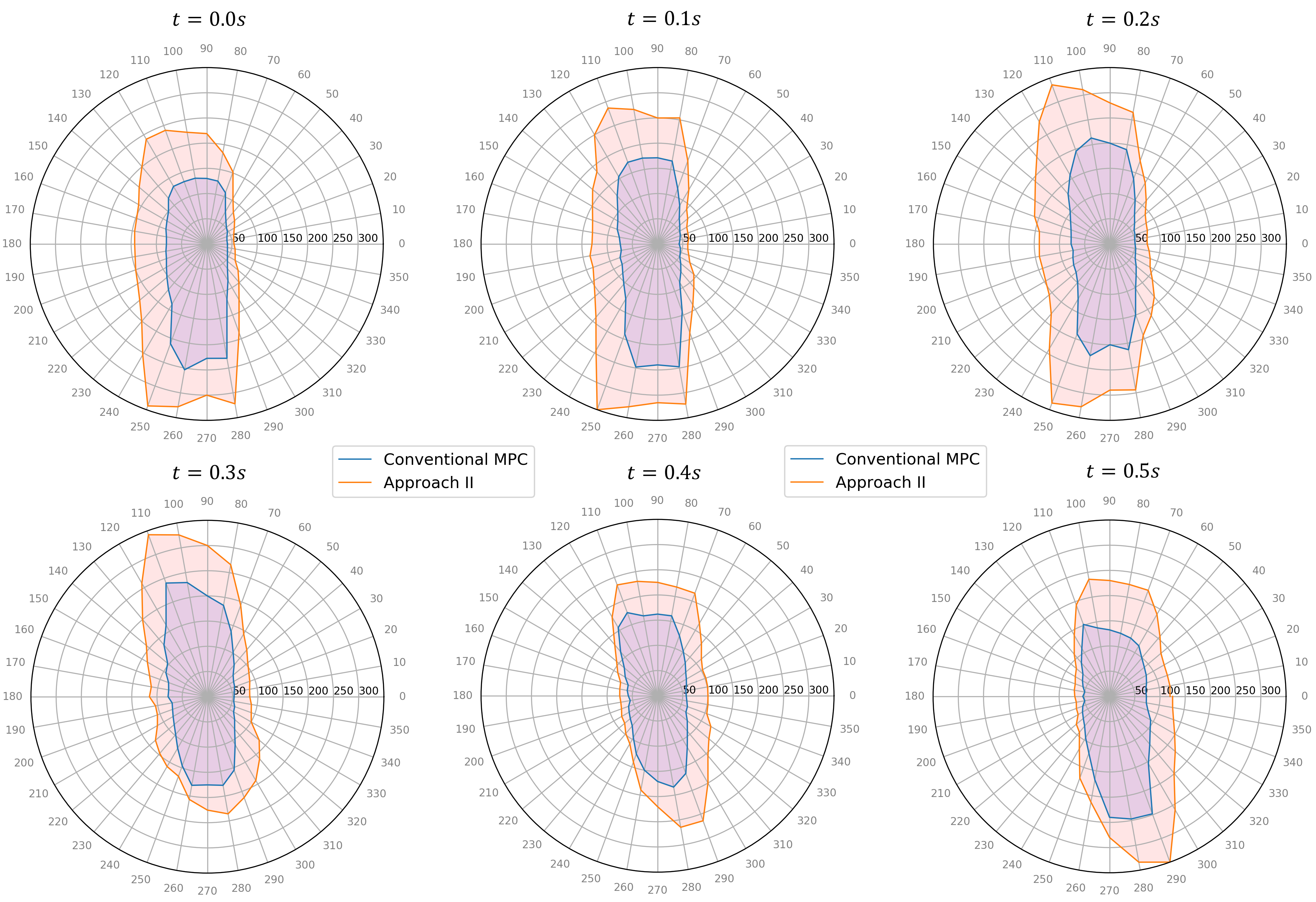}
\vspace{-2mm}
\caption{Comparison between conventional MPC (without projection) and \emph{Approach II} (with projection) in push recovery. $\theta = \ang{90}$ corresponds to a forward push, while $\theta = \ang{0}$ and $\theta = \ang{180}$ represent pushes to
the right and left directions, respectively. The radius shows the magnitude in $N$ of the pushes applied for $\Delta t = 0.2$ s uniformly that the controller was able to reject. In all cases, \emph{Approach II} performed better.}
\label{fig:s1c4-push-recovery-comparison}
\end{figure*}
\end{center}

To further investigate the performance of our proposed projection \emph{Approach II}, we apply every 0.1 s some random disturbances in the range $c_{x,y} \in (0,0.5)$, and $\dot c_{x,y} \in (0, 0.1)$ respectively, from $t=5.0$ s to $t=6.0$ s. Figure \ref{fig:s1c1_2} (right) shows the DCM position and Fig.~\ref{fig:s1c1_2} (left) is the resulting walking velocity for this case.
It is obvious that \emph{Approach II} is able to achieve the task (tracking the desired velocity $v = 1$ m/s) even in presence of random disturbances by projecting the measured state. 

To show how our projection can result in a robust walking for the full humanoid robot, we perform a full body simulation where we use the following weights for the slow MPC with projection based on \emph{Approach II} with $\alpha_x=\alpha_y=1,\beta_x=\beta_y=100, \delta_{x}=5, \delta_{y}=20$ and $\eta_x=\eta_y=0$. $\beta_{x},\beta_{y}$ is chosen to be 100 to bring the reference ZMP close to the middle of the stance foot. $\delta_{x}$ is 5 to reduce the reference velocity slightly in order to make the walking pattern more robust against the disturbances and $\delta_{y}$ is 20 to force the resulting step width to be equal to the pelvis length. This prevents the violation of the minimum step width constraint, which can happen when the swing foot is close to touching the ground. In this simulation, we exert four lateral pushes $F_d=-35N$, $F_d=45$ N, $F_d=-60$ N, and $F_d=70$ N uniformly to the robot at $t=4.2$ s, $t=8.4$ s, $t=13.2$ s, and $t=16.2$ s respectively, during $\Delta t = 0.2$ s. Without projection, although the robot rejects the first lateral push, it falls after the second push. This is because the DCM starts to diverge rapidly after the second push at $t = 8.4$ s and begins to move away from the boundaries of the viability kernel around $t = 9.3$ s (Fig.~\ref{fig:s1c2} right). In contrast, by using \emph{Approach II}, the robot is able to reject all the external pushes because the initial measured state is modified so that the state of the nominal model remains inside the viability kernel and the MPC problem remains feasible. Therefore, the robot can preserve its balance and successfully achieve the task (Fig.~\ref{fig:s1c2} left). 

To systematically show the effectiveness of the proposed projection, we performed extensive simulation tests with different external uniform forces applied to the robot in every direction (from $\ang{0}$ to $\ang{360}$, every $\ang{10}$) and every time step of a complete footstep (every $0.1 $ s of $0.5$ s) during $\Delta t = 0.2$ s. The cost weights are $\alpha_x=\alpha_y=1,\beta_x=\beta_y=100, \delta_x=\delta_y=20$, and $\eta_x=\eta_y=0$.
Note that changing the cost weights affects the amount of push that can be rejected, and here we only aim to show the difference for one set of cost weights. As we can see clearly in Fig.~\ref{fig:s1c4-push-recovery-comparison}, the proposed projection increases the robustness of the controller against external pushes in all directions and all time instances in one step.

\begin{center}
\begin{figure*}[!tbp] 
\includegraphics[width = .95\textwidth]{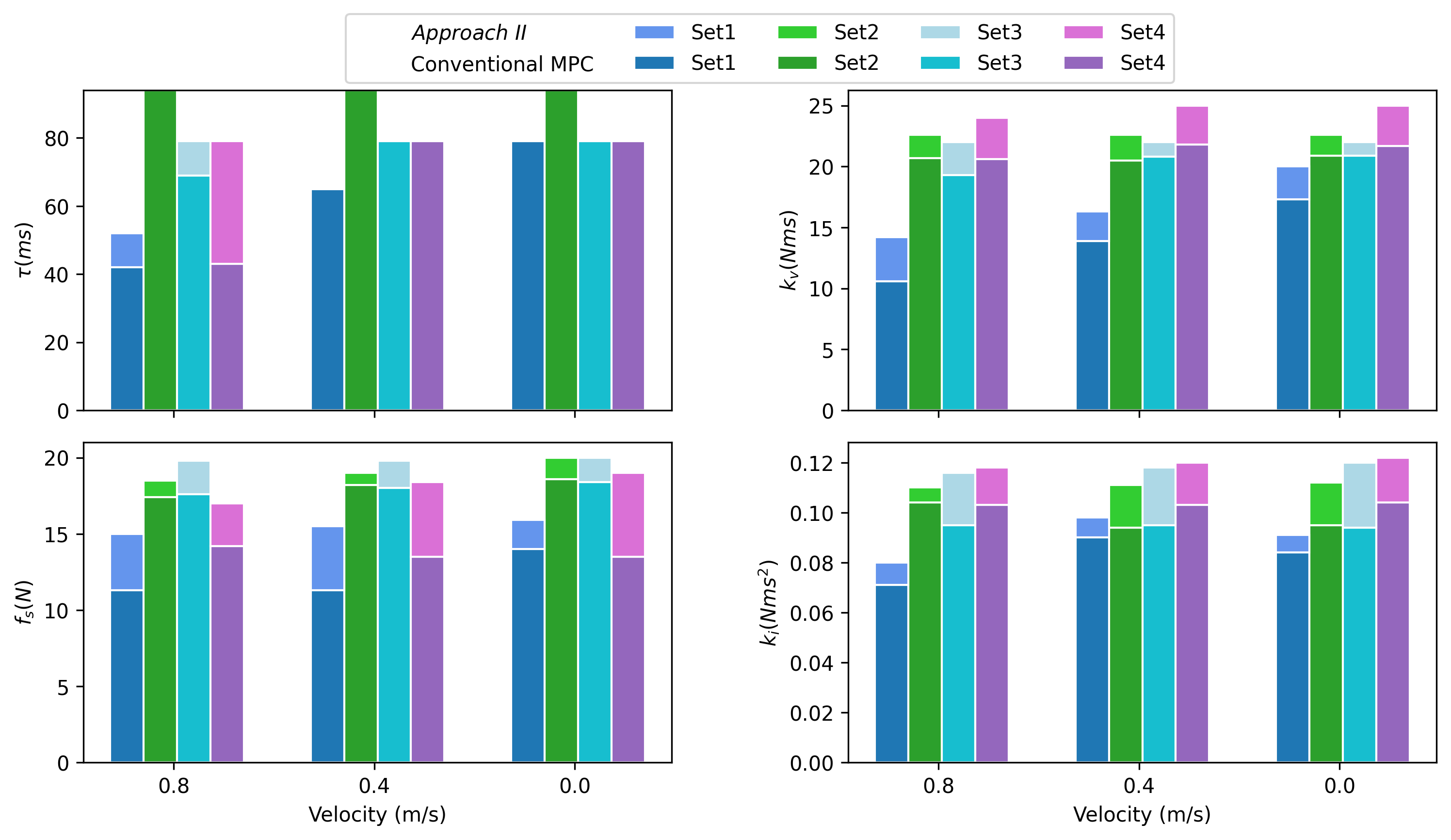}
\vspace{-2mm}
\caption{Maximum tolerable time delay ($\tau$), viscous friction factor ($k_v$), Coulomb friction ($f_s$), and rotor inertia factor ($k_i$) for different velocity of walking and various MPC cost weights. The sets of cost weights are as follows: Set1: $\alpha_x = \alpha_y = 1, \beta_x = \beta_y = \delta_x = \delta_y = \eta_x = \eta_y = 0$, Set2: $\alpha_x = \alpha_y = 1, \beta_x = \beta_y = \delta_x = \delta_y = \eta_x = \eta_y = 100$, Set3: $\alpha_x = \alpha_y = 1, \beta_x = \beta_y = \delta_x = \delta_y = \eta_x = \eta_y = 500$, Set4: $\alpha_x = \alpha_y = 100, \beta_x = \beta_y = \delta_x = \delta_y = \eta_x = \eta_y = 1000$. The result of Conventional MPC without projection is also included for each set with similar color}
\label{fig:s2c1-max-delay-actuator-properties}
\end{figure*}
\end{center}

\subsection{Effect of different realistic uncertainties}\label{sec:res,uncertainty}

In this subsection, we study the effects of different realistic uncertainties on performance. We perform these tests to quantify the robustness of the proposed two-level MPC controller with viability projection and to motivate how changing the slow MPC cost weights can add extra robustness to the control pipeline. We also compare the results of Conventional MPC without viability projection with our proposed approach to show how much robustness our method can add to the control pipeline in the presence of different uncertainties. Then, in the next subsection, we use BO to systematically find the optimal cost weights that trade-off performance and robustness.   

\subsubsection{Computational delay}
Solving an iLQG problem rather than a traditional instantaneous inverse dynamics problem (a QP) for the whole-body controller comes at the price of a higher computational cost \cite{koenemann2015whole}. Here we try to quantify how this delay would affect the performance of our control pipeline. We carry out walking simulations with three different velocities, i.e. $v = 0, \, 0.4, \, 0.8$ m/s. We include a constant time delay in our simulations to explore the maximum delay that the robot can tolerate without falling down (Fig.~\ref{fig:s2c1-max-delay-actuator-properties}), and also to examine the effect of the slow MPC cost weights on this maximum time delay for various walking velocities. We have taken this computational delay $\tau$ into account in our simulation by applying the predicted control input based on the measurement at $t_i-\tau$ for the current time $t_i$.

As expected, larger walking velocities need faster control loops (Fig.~\ref{fig:s2c1-max-delay-actuator-properties}(top,left)). We can also observe that the maximum time delay depends on the slow MPC cost weights; therefore, finding proper cost weights can increase the robustness to computational delay. For some sets of weights (Set2 and Set3 of Fig.~\ref{fig:s2c1-max-delay-actuator-properties}) the maximum time delay is the same for all the velocities because the weights result in walking velocities close to zero. Also, selecting higher cost weights does not always guarantee further robustness (comparing the results of Set2 and Set3 in Fig.~\ref{fig:s2c1-max-delay-actuator-properties}). It is interesting also to observe that Approach II does not improve robustness with respect to computational delay at low and medium walking velocities ($v = 0, \, 0.4$ m/s).
Based on the results reported for a state-of-the-art solver \cite{mastalli2019crocoddyl}, we would have roughly a computational delay of 25 ms which, based on our analyses, seems to be tolerable for real robot experiments. 

\subsubsection{Actuator uncertainties}
To make our simulations more realistic, we assume that the robot actuators are not perfect and do not deliver the desired torques computed by iLQG. We consider the main effects present in electric actuators with gearboxes, i.e. viscous friction $k_v\dot{q}$, Coulomb friction $f_s$, and rotor inertia effects $k_i\ddot{q}$. The torques applied to the joints ($\tau_a$) are computed as follows
\begin{align}
\label{eq:motor-torques}
\tau_{a} = \tau_{iLQG} - k_i\ddot{q} - k_v\dot{q} - f_s \, sgn(\dot{q}),
\end{align} 
where $\ddot{q}$ and $\dot{q}$ are the joint acceleration and velocity vectors, respectively. $k_i, k_v, f_s$ are constant values depending on the actuators, and $sgn()$ stands for the sign function.
Figure~\ref{fig:s2c1-max-delay-actuator-properties} reports the maximum admissible values of $k_i, k_v, f_s$ individually for the same set of cost weights used in the previous section. Again, as we move from right to left in each subplot of Fig.~\ref{fig:s2c1-max-delay-actuator-properties} (larger walking velocities), we can see a decrease in robustness to actuator uncertainty. Interestingly, by changing the cost weights, we might increase robustness to some uncertainties, while decreasing it against other uncertainties. For instance, for a walking velocity of 0.8 m/s, when $\alpha=1$ and other cost terms are zero (Set1), we can see the least robustness against different uncertainties. By increasing $\beta,\delta,\eta$, we see that the robustness against all uncertainties is increased; however, the maximum robustness for different uncertainties is achieved with different cost weights (compare different Sets with each other). This again shows that the cost weights of the slow MPC are crucial for robustness, but finding them is not trivial. Finally, as expected, there is an increase in robustness in cases where Approach II is used compared to the Conventional MPC without any projection.

\subsection{Using BO to find optimal cost weights}\label{sec:res,BO}
In this subsection we demonstrate the application of BO (Section \ref{sec:BO}) to generate robust gaits in the presence of external disturbances and different uncertainties (i.e. computational delay and actuator imperfection). The slow MPC cost weights (i.e. $\alpha, \beta, \delta$, and $\eta$) are the decision variables. The  weights can vary from 0 to 1000. In all simulation episodes, we consider a constant computational delay $\tau = 25$ ms, which is based on the results reported in \cite{mastalli2019crocoddyl}. We set the unknown actuator parameters $f_{s} = 1.5$ N, $k_i = 0.005$ Nms$^2$, and $k_v = 1.5$ Nms. For each new set of weights that BO proposes for each episode, we carried out 50 simulations with random external disturbances, which are decomposed into two simultaneous external forces in sagittal and lateral directions, applied at random times during a complete footstep (the six cases in Fig.~\ref{fig:s1c4-push-recovery-comparison}) for $\Delta t = 0.2$ s and used the average cost as the cost of this BO query. Note that we first sampled 50 random disturbances inside $(-45, 35)$ N for lateral direction and inside $(-80, 110)$ N for sagittal direction and applied the same disturbance for each query to make sure that the cost of any set of decision variables remains the same if recomputed.

\begin{table*}[!tbp]
\centering
\caption{Selecting slow MPC cost weights manually in the presence of external disturbances and different uncertainties. 'No. of Falls' shows the cases where robot fell down when subjected to 50 random external pushes. 'Performance' is the average velocity tracking cost over 50 episodes.}
\begin{tabular}{@{}cccccccccccccccccc@{}}
\begin{tabular}[c]{@{}c@{}}Set of Cost \\ weights\end{tabular} & 1    & 2    & 3    & 4    & 5   & \textbf{6}    & \textbf{7}    & 8    & 9    & 10   & 11   & 12   & 13   & 14   & 15   & 16   & \textbf{\begin{tabular}[c]{@{}c@{}}BO\\ Result\end{tabular}} \\ \midrule
$\alpha_x$                                                       & 1.0  & 1.0  & 1.0  & 1.0  & 1.0 & \textbf{1.0}  & 1.0           & 1.0  & 10   & 10   & 10   & 100  & 100  & 10   & 200  & 250  & \textbf{999}     \\
$\alpha_y$                                                       & 1.0  & 1.0  & 1.0  & 1.0  & 1.0 & \textbf{1.0}  & \textbf{1.0}  & 1.0  & 10   & 10   & 1.0  & 1000 & 50   & 10   & 10   & 250  & \textbf{1.0}     \\
$\beta_x$                                                        & 1000 & 500  & 100  & 50   & 10  & \textbf{100}  & \textbf{100}  & 100  & 100  & 1000 & 100  & 700  & 100  & 100  & 500  & 700  & \textbf{832.61}  \\
$\beta_y$                                                        & 1000 & 500  & 100  & 50   & 10  & \textbf{100}  & \textbf{100}  & 100  & 100  & 1000 & 100  & 700  & 100  & 100  & 500  & 700  & \textbf{716.09}  \\
$\delta_x$                                                       & 1000 & 500  & 100  & 50   & 10  & \textbf{50}   & \textbf{30}   & 10   & 30   & 30   & 30   & 500  & 50   & 50   & 500  & 500  & \textbf{133.95}  \\
$\delta_y$                                                       & 1000 & 500  & 100  & 50   & 10  & \textbf{50}   & \textbf{30}   & 30   & 30   & 30   & 30   & 500  & 50   & 50   & 50   & 100  & \textbf{311.70}  \\
$\eta_x$                                                         & 1000 & 500  & 100  & 50   & 10  & \textbf{1000} & \textbf{1000} & 1000 & 1000 & 1000 & 1000 & 100  & 1000 & 1000 & 1000 & 1000 & \textbf{0.0}     \\
$\eta_y$                                                         & 1000 & 500  & 100  & 50   & 10  & \textbf{1000} & \textbf{1000} & 1000 & 1000 & 1000 & 1000 & 500  & 1000 & 1000 & 1000 & 1000 & \textbf{1000}    \\ \midrule
No. of Falls                                                   & 3    & 5    & 2    & 3    & 12  & \textbf{0}    & \textbf{0}    & 3    & 13   & 6    & 6    & 22   & 9    & 11   & 7    & 23   & \textbf{0}       \\ \midrule
Performance                                                    & 38.2 & 35.6 & 29.4 & 23.8 & 7.5 & \textbf{25.3} & \textbf{18.5} & 11.7 & 3.6  & 6.8  & 4.3  & 4.2  & 2.1  & 4.7  & 3.3  & 2.6  & \textbf{5.1}    
\end{tabular}
\label{tab:BO-manual-cost-weight}
\end{table*}

\begin{figure}[!tbp] 
\includegraphics[width = \columnwidth]{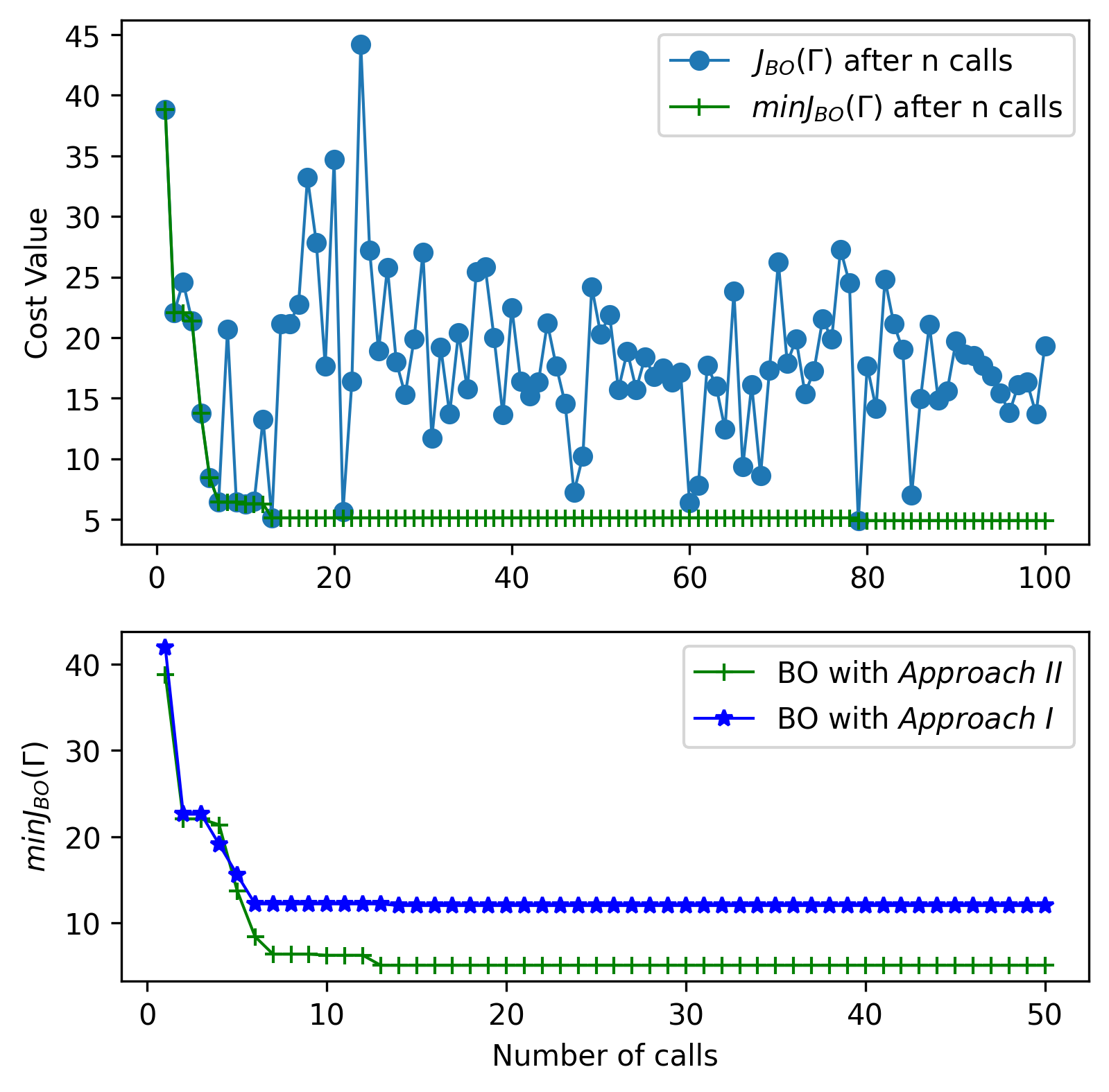}
\vspace{-5mm}
\caption{(top) BO search history with random external disturbances using Approach II. Note that BO could find a very good solution in 13th iteration after which little improvement in the cost value is observed. (bottom) comparison between Approach I and Approach II when they were used with BO. Approach II outperform Approach I in the same condition of external disturbances and different uncertainties}
\label{random_disturbances}
\end{figure}

We have initialized BO with $\alpha_x=\alpha_y=1$ and the other cost weights set to 1000. Although BO found the optimal cost weights $\alpha_x = 470, \alpha_y = 20, \beta_x = 319, \beta_y = 900, \delta_x =  443, \delta_y = 81, \eta_x = 488$ and $\eta_y = 489$ after 79 queries, the cost have been settled after 13 iterations. For this set of decent cost weights $\alpha_x = 999, \alpha_y = 1, \beta_x = 833, \beta_y = 716, \delta_x =  134, \delta_y = 312, \eta_x = 0, \eta_y = 1000$ after 13 queries, we observed that the robot can reject all 50 disturbances in the presence of the computational delay and the actuators uncertainties. Fig.~\ref{random_disturbances} (top) shows the evolution of cost values over all iterations and the minimum cost value among all BO calls until the current iteration. Importantly, the fact that BO could find a robust policy against different uncertainties and external disturbances within a few iterations suggests that such trial and error procedure could directly be done on a real robot without any prior knowledge on the uncertainties.

We also repeated the same experiment using Approach I, where the new parameters $\eta_x$ and $\eta_y$ could vary from 0 to 100000. Figure~\ref{random_disturbances} (bottom) shows that our proposed approach (Approach II) performs better than Approach I in the presence of the same uncertainties. The optimal cost weights for BO with Approach I are $\alpha_x = 1000, \alpha_y = 108, \beta_x = 355, \beta_y = 690, \delta_x =  355, \delta_y = 968, \eta_x = 508, \eta_y = 978, \theta_x = 13560, \theta_y = 90721$, which are found after 14 queries. Interestingly, BO found large values for $\theta$ which confirms our observation that we need large $\theta$ for a successful walking. Although using these cost weights the robot is able to track the desired velocity quite well, it falls down 9 times out of the 50 episodes with random external pushes.

Finally, to demonstrate how challenging it is to find the slow MPC cost weights based on expert knowledge, we tried to find the optimal cost weights for this experiment using our intuition and compared the results with BO in Table~\ref{tab:BO-manual-cost-weight}. First, we started from a set of cost weights that tries to move away the solutions from the boundaries of all constraints (column 1 of Table~\ref{tab:BO-manual-cost-weight}) and observed that not only the velocity tracking is poor, but the robot is also not able to reject all disturbances and fails to keep balance 3 times out of 50 random disturbances. Then, we started to decrease the cost weights to see how it affects the robot performance (Columns 2 to 5) and observed that the robot can track the velocity better but still it falls down a few times. After 6 trial, we found a set of cost weights that the robot is able to walk robustly without falling down (Table~\ref{tab:BO-manual-cost-weight}), but the velocity tracking still needs to be improved. Our next efforts to increase the performance while avoiding a fall were not successful as shown in Table~\ref{tab:BO-manual-cost-weight}. Comparing this procedure with BO shows that 1) it is far from trivial to find good cost weights and 2) BO can find optimal parameters without any expert knowledge better than the expert. 

\section{CONCLUSIONS AND FUTURE WORK}\label{sec:conclusion}
In this paper, we presented a two-level MPC framework for generating robust walking for humanoid robots. In this approach, the higher level (slow MPC) uses LIPM abstraction of the robot dynamics and constructs a linear MPC problem to track a desired velocity. To guarantee that the trajectories from the slow MPC do not diverge, we proposed a novel viability-based projection. The trajectories generated by the higher level MPC (slow MPC) are then tracked by a low-level MPC (fast MPC), which is based on iLQG. We showed the effectiveness of this framework in dealing with different sets of realistic uncertainties, i.e. external disturbances, unmodeled actuator dynamics and computational delay of the fast MPC. Finally, we presented a systematic approach based on Bayesian Optimization to find the best cost weights of the slow MPC that trades off performance and robustness.

It is important to note that depending on the full capability of the robot, at some point failure is unavoidable. In fact, even though using our proposed approach we can all the time project the measured state inside the viability kernel of the simplified model, in some states the fast MPC fails to track the desired trajectories from the slow MPC. In this case, if we know that we are in a state that is impossible to recover from (the full robot viability kernel), we can switch to a safe-fall mode that minimizes damage to the robot hardware. While computing the viability kernel for the full robot is extremely hard, one can try to learn this kernel for a specific robot and a given control policy (for instance the optimal controller obtained from BO) by sampling different initial states and rolling out the robot motion in the simulation environment and learning a bounded set (with a desired shape) that contains only the viable states. This can be seen as a potential research direction for future work.
Another  interesting  extension of this work is  to  use  BO  to  modify  simultaneously  both the slow and fast MPC cost weights.  Finally, we intend to test the proposed control framework on a real humanoid or biped robot.

\section*{APPENDIX A}
LIPM dynamics in the sagittal direction can be formulated as
\begin{align}
\label{eq_app:LIPM}
 \ddot c_x = \omega_0^2(c_x-z_{x})
\end{align}
where $z_{x} \in [x^f \pm \frac{L_f}{2}]$. Considering the CoM $c_x$ and DCM $\xi_x=c_x+\dot c_x/\omega_0$ as the state variables, the equations in state space can be written as
\begin{subequations}
    \begin{align}
    \label{eq_app:LIPM-stable-part}
 \dot c_x = \omega_0(\xi_x-c_x) \\
 \label{eq_app:LIPM-unstable-part}
 \dot{\xi_x} = \omega_0(\xi_x-z_x) 
    \end{align}
\end{subequations}
Solving \eqref{eq_app:LIPM-unstable-part} as a final value problem (with fixed $z_x$), we have
\begin{align}
\label{eq_app:DCM-forward}
  \xi_{T,x}= (\xi_{t,x}-z_x)e^{\omega_0 (T_s-t)} + z_x , \quad 0 \leq t < T_s
\end{align}
Defining the DCM offset of current and next step as $b_{t,x} = \xi_{t,x} - x^f_0$ and $b_{T,x} = \xi_{T,x} - x^f_1$, we have
\begin{align}
\label{eq_app:DCM-offset}
  b_{T,x} + x^f_1= \left(b_{t,x} + x^f_0 -z_x\right) e^{\omega_0(T_s-t)}+ z_x
\end{align}
Considering the ZMP on the foot edge for computing maximum DCM offset $z_x = x^f_0 + \frac{L_f}{2} $, we have 
\begin{align}
\label{eq_app:DCM-offset_max}
  b_{T,x} - \frac{L_f}{2} + x^f_1= \left(b_{t,x} -\frac{L_f}{2}\right) e^{\omega_0(T_s-t)}+ x^f_0
\end{align}

\section*{APPENDIX B}

We describe here the computation of the viability kernel in lateral directions based on the coordinate system in Fig.~\ref{fig:DCM}.

\subsection{Lateral outward direction}
Without loss of generality, we assume that the right foot is in stance in the current step. Considering the swing foot velocity constraint in lateral outward direction
\begin{align}
\label{eq_app:swing_max_y_right_outward}
    \overline{y}_1^f = y_s^f + \overline{v}_y (t_{td}-t)
\end{align}

Combining this with the maximum step width constraint in the current step, we compute the allowable foot landing location for the outward direction in the current step as

\begin{align}
\label{eq_app:foot_max_right_outward_y}
    y_1^{f,rea,in} =\; &(y_1^f - y_0^f)_{max} \nonumber\\
    =\; &\text{min } [\overline{y}_1^f - y_0^f,(L_p+\overline{W})]
\end{align}

We can write down the DCM time evolution in the current step as
\begin{align}
\label{eq_app:DCM-current-right-outward}
    b_{T,y,l} - \frac{W_f}{2} + y_1^f= \left (b_{t,y,r} - \frac{W_f}{2}\right) e^{\omega_0(T_s-t)}+ y_0^f
\end{align}

According to Fig.~\ref{fig:DCM} (right), for the next two steps we can write the DCM equation as
\begin{subequations}
\label{eq_app:DCM-current-left-outward}
    \begin{align}
    -(L_p+\underline{W})= \; &\left(\overline{b}_{T,y,l}-\frac{W_f}{2}\right)e^{\omega_0T}
    -\left(\overline{b}_{T,y,r}- \frac{W_f}{2}\right)\\
    (L_p+\overline{W})= \; &\left(\overline{b}_{T,y,r}-\frac{W_f}{2}\right)e^{\omega_0T}
    -\left(\overline{b}_{T,y,l}- \frac{W_f}{2}\right)
    \end{align}
\end{subequations}

Using \eqref{eq_app:DCM-current-left-outward}, we compute $\overline{b}_{T,y,l}$
\begin{align}
    \overline{b}_{T,y,l} &= \frac{W_f}{2}
    -\frac{L_{p}}{1+e^{\omega_0 T_s}}
    -\frac{\overline{W}-\underline{W}e^{\omega_0 T_s}}{1-e^{2\omega_0 T_s}}
\end{align}

Substituting this equation and \eqref{eq_app:foot_max_right_outward_y} into \eqref{eq_app:DCM-current-right-outward} we compute the viability kernel boundary in lateral outward direction as
\begin{align}
\label{eq_app:DCM-offset-max-y-right-outward}
    \overline{b}_{t,y,r,in} &= \frac{W_f}{2}
    +\bigg[y_1^{f,rea,in}
    -\frac{L_{p}}{1+e^{\omega_0 T_s}} \nonumber\\
    &-\frac{\overline{W}-\underline{W}e^{\omega_0 T_s}}{1-e^{2\omega_0 T_s}}\bigg]
    e^{-\omega_0(T_s-t)}
\end{align}

For the case in which the left foot is in stance we have
\begin{align}
\label{eq_app:swing_max_y_left_outward}
    &\overline{y}_1^f = y_s^f - \overline{v}_y (t_{td}-t)
\end{align}

and
\begin{align}
\label{eq_app:foot_max_left_outward_y}
    y_1^{f,rea,in} =\; &(y_1^f - y_0^f)_{max} \nonumber\\
    =\; &\text{max } [\overline{y}_1^f - y_0^f,-(L_p+\overline{W})]
\end{align}

With the same procedure we obtain
\begin{align}
\label{eq_app:DCM-offset-max-y-left-outward}
    \overline{b}_{t,y,l,in} &= -\frac{W_f}{2}
    + \bigg[y_1^{f,rea,in}
    +\frac{L_{p}}{1+e^{\omega_0 T_s}} \nonumber\\
    &+\frac{\overline{W}-\underline{W}e^{\omega_0 T_s}}{1-e^{2\omega_0 T_s}}\bigg]
    e^{-\omega_0(T_s-t)}
\end{align}

\subsection{Lateral inward direction}
Again we assume that the right foot is in stance in the current step. Considering the swing foot velocity constraint in lateral inward direction
\begin{align}
\label{eq_app:swing_max_y_right_inward}
    \underline{y}_1^f = y_s^f - \overline{v}_y (t_{td}-t)
\end{align}

Combining this with the minimum step width constraint in the current step, we compute the allowable foot landing location for the inward direction in the current step as

\begin{align}
\label{eq_app:foot_max_right_inward_y}
    y_1^{f,rea,out} =\; &(y_1^f - y_0^f)_{min} \nonumber\\
    =\; &\text{max } [\underline{y}_1^f - y_0^f,(L_p+\underline{W})]
\end{align}

We can write down the DCM time evolution in the current step as
\begin{align}
\label{eq_app:DCM-current-right-inward}
    b_{T,y,l} + \frac{W_f}{2} + y_1^f= \left(b_{t,y,r} + \frac{W_f}{2}\right) e^{\omega_0(T_s-t)}+ y_0^f
\end{align}

According to Fig.~\ref{fig:DCM} (left), for the next two steps we can write the DCM equation as
\begin{subequations}
\label{eq_app:DCM-current-left-inward}
    \begin{align}
    -(L_p+\overline{W})= \; &\left(\overline{b}_{T,y,l}+\frac{W_f}{2}\right)e^{\omega_0T} 
    -\left(\overline{b}_{T,y,r}+ \frac{W_f}{2}\right)\\
    (L_p+\underline{W})= \; &\left(\overline{b}_{T,y,r}+\frac{W_f}{2}\right)e^{\omega_0T}
    -\left(\overline{b}_{T,y,l}+ \frac{W_f}{2}\right)
    \end{align}
\end{subequations}

Using \eqref{eq_app:DCM-current-left-inward}, we compute $\overline{b}_{T,y,l}$
\begin{align}
    \overline{b}_{T,y,l} &= -\frac{W_f}{2}
    -\frac{L_{p}}{1+e^{\omega_0 T_s}}
    -\frac{\underline{W}-\overline{W}e^{\omega_0 T_s}}{1-e^{2\omega_0 T_s}}
\end{align}

Substituting this equation and \eqref{eq_app:foot_max_right_inward_y} into \eqref{eq_app:DCM-current-right-inward} we compute the viability kernel boundary in lateral inward direction as
\begin{align}
\label{eq_app:DCM-offset-max-y-right-inward}
    \overline{b}_{t,y,r,out} &= -\frac{W_f}{2}
    +\bigg[y_1^{f,rea,out}
    -\frac{L_{p}}{1+e^{\omega_0 T_s}} \nonumber\\
    &-\frac{\underline{W}-\overline{W}e^{\omega_0 T_s}}{1-e^{2\omega_0 T_s}}\bigg]
    e^{-\omega_0(T_s-t)}
\end{align}

For the case in which the left foot is in stance we have
\begin{align}
\label{eq_app:swing_max_y_left_inward}
    &\underline{y}_1^f = y_s^f + \overline{v}_y (t_{td}-t)
\end{align}
and
\begin{align}
\label{eq_app:foot_max_left_inward_y}
    y_1^{f,rea,out} =\; &(y_1^f - y_0^f)_{min} \nonumber\\
    =\; &\text{min } [\underline{y}_1^f - y_0^f,-(L_p+\underline{W})]
\end{align}
With the same procedure we obtain
\begin{align}
\label{eq_app:DCM-offset-max-y-left-inward}
    \overline{b}_{t,y,l,out}  &= \frac{W_f}{2}
    + \bigg[y_1^{f,rea,in}
    +\frac{L_{p}}{1+e^{\omega_0 T_s}} \nonumber\\
    &+\frac{\underline{W}-\overline{W}e^{\omega_0 T_s}}{1-e^{2\omega_0 T_s}}\bigg]
    e^{-\omega_0(T_s-t)}
\end{align} 

\section*{ACKNOWLEDGMENT}
We would like to thank Ahmad Gazar for fruitful discussions on RMPC and SMPC, and Jia-Jie Zhu for discussions on black-box optimization algorithms and corresponding software.

\bibliography{Master}
\bibliographystyle{IEEEtran}

\addtolength{\textheight}{-12cm}   




\begin{IEEEbiography}[{\includegraphics[width=1in,height=1.25in,clip,keepaspectratio]{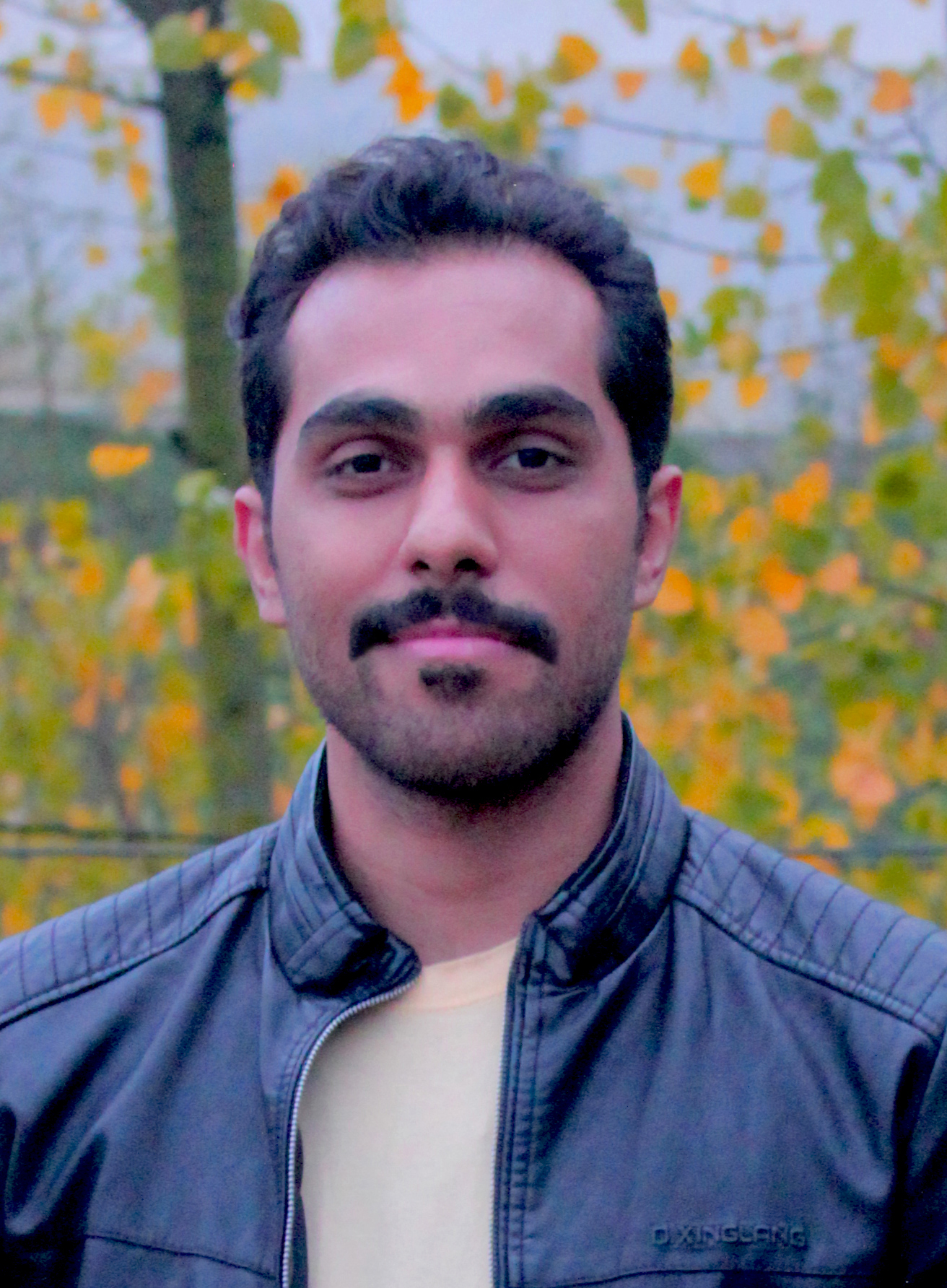}}]{Mohammad Hasan Yeganegi}
received his BSc in Mechanical Engineering from Bu-Ali Sina University, Hamedan, Iran in 2016, and his MSc degree in Mechanical Engineering from K. N. Toosi University of Technology (KNTU), Tehran, Iran in 2019. From October 2019 to January 2021, he did an (virtual) internship at the Movement Generation and Control Group, Max-Planck Institute for Intelligent Systems, T\"ubingen, Germany. His main research interest is legged locomotion control.
\end{IEEEbiography}
\begin{IEEEbiography}[{\includegraphics[width=1in,height=1.25in,clip,keepaspectratio]{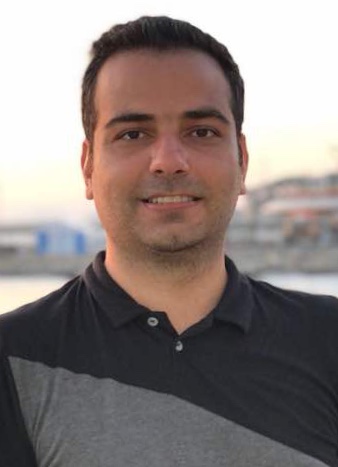}}]{Majid Khadiv}
is a postdoctoral researcher at the Movement Generation and Control Group, Max-Planck Institute for Intelligent Systems. He received his BSc degree in Mechanical Engineering from Isfahan University of Technology (IUT) in 2010, and his MSc and PhD degrees in Mechanical Engineering from K. N. Toosi University of Technology, Tehran, Iran in 2012 and 2017. Majid joined the Iranian national humanoid project, Surena III, and worked as the head of dynamics and control group from 2012 to 2015. He also spent one-year of his PhD as a visiting researcher at the Max-Planck Institute for Intelligent Systems. His main research interest is control of robots with contact interaction.
\end{IEEEbiography}

\begin{IEEEbiography}[{\includegraphics[width=1in,height=1.25in,clip,keepaspectratio]{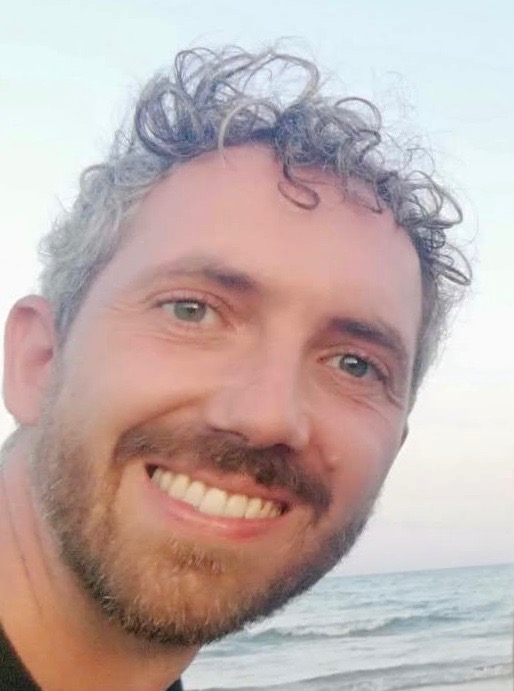}}]{Andrea Del Prete} is an Assistant Professor in the Industrial Engineering Department at the University of Trento (Italy). He received the degree (Hons.) in computer engineering from the University of Bologna (Italy) in 2009, and the Ph.D. degree in robotics from the Cognitive Humanoids Laboratory (Italian Institute of Technology, Genova, Italy) in 2013.
From 2014 to 2017 he was a Postdoctoral Researcher with the Laboratoire d’Analyse et d’Architecture des Systemes (CNRS, Toulouse, France). In 2018, he had worked as Senior Researcher at the Max Planck Institute for Intelligent Systems (Tübingen, Germany). His research focuses on the use of optimization algorithms for control, planning and estimation of autonomous robots, with a special focus on legged locomotion.
\end{IEEEbiography}

\begin{IEEEbiography}[{\includegraphics[width=1in,height=1.25in,clip,keepaspectratio]{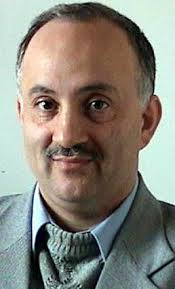}}]{S. Ali. A. Moosavian}
received his B.Sc. degree in 1986 from Sharif University of Technology and the M.Sc. degree in 1990 from  Tarbiat  Modaress  University  (both  in  Tehran),  and  his Ph.D.   degree   in   1996   from   McGill   University   (Montreal, Canada),  all  in  Mechanical  Engineering.  He  is  a  Professor with  the  Mechanical  Engineering  Department  at  K.  N.  Toosi University  of  Technology  (KNTU)  in  Tehran  since  1997. His research interests are in the areas of dynamics modeling and motion/impedance control of terrestrial, legged and space robotic systems. He has published more than 200 articles in peer-reviewed journals and conference proceedings. He is one of the Founders of the ARAS Research Group, and the Manager of Center of Excellence in Robotics and Control at KNTU.
\end{IEEEbiography}

\begin{IEEEbiography}[{\includegraphics[width=1in,height=1.25in,clip,keepaspectratio]{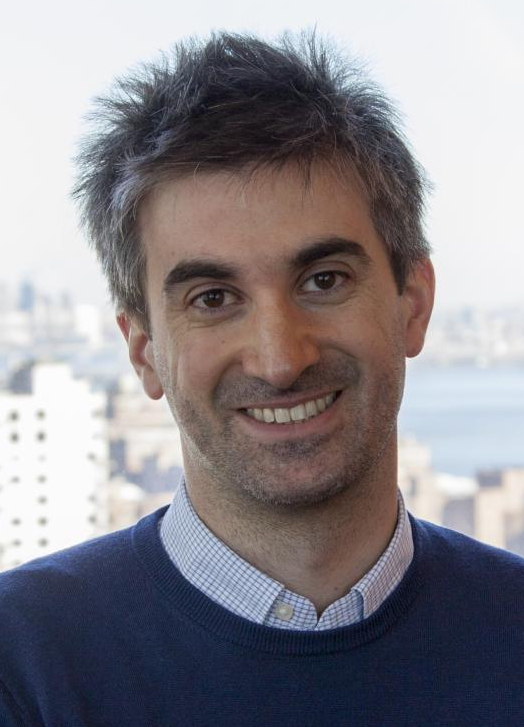}}]{Ludovic Righetti}
is an Associate Professor in the Electrical and Computer Engineering Department and in the Mechanical and Aerospace Engineering Department at the Tandon School of Engineering at New York University and a Senior Researcher at the Max-Planck Institute for Intelligent Systems in T\"ubingen, Germany. He holds an engineering diploma in Computer Science and a Doctorate in Science from the Ecole Polytechnique F\'ed\'erale de Lausanne (Switzerland). His research focuses on the planning and control of movements for autonomous robots, with a special emphasis on legged locomotion and manipulation.

\end{IEEEbiography}

\end{document}